\documentclass[lettersize,journal]{IEEEtran}
\usepackage{amsmath,amsfonts}
\usepackage{algorithmic}
\usepackage{algorithm}
\usepackage{array}
\usepackage[caption=false,font=normalsize,labelfont=sf,textfont=sf]{subfig}
\usepackage{textcomp}
\usepackage{stfloats}
\usepackage{url}
\usepackage{verbatim}
\usepackage{graphicx}
\usepackage{cite}
\usepackage{color}
\usepackage{booktabs}
\usepackage{multirow}
\usepackage[T1]{fontenc}    
\usepackage[numbers]{natbib}
\usepackage{hyperref}
\usepackage[capitalize,noabbrev]{cleveref}
\hypersetup{
	colorlinks=true,
	linkcolor=blue,
	citecolor=blue,
}

\DeclareMathOperator{\Gini}{Gini}
\DeclareMathOperator{\Flatten}{Flatten}

\begin{document}
	
	\title{Releasing Inequality Phenomenon in $\ell_{\infty}$-norm Adversarial Training via Input Gradient Distillation}
	
	\author{Junxi~Chen,
		Junhao~Dong, 
		Xiaohua~Xie,~\IEEEmembership{Member,~IEEE},
		and~Jianhuang~Lai,~\IEEEmembership{Senior~Member,~IEEE}
		\thanks{Manuscript received XXX XX, XXXX; revised XXX XX, XXXX. This project is supported by National Natural Science Foundation of China (12326618) and the Project of Guangdong Provincial Key Laboratory of Information Security Technology (2023B1212060026).} \thanks{Corresponding author: Xiaohua Xie.}
		\thanks{The authors are with the School of Computer Science and Engineering, Sun Yat-sen University, 510006, Guangzhou, China, and with the Guangdong Province Key Laboratory of Information Security Technology, 510006, Guangzhou, China, and also with and the Key Laboratory of Machine Intelligence and Advanced Computing, Ministry of Education, 510006, Guangzhou, China.(e-mail: chenjx353@mail2.sysu.edu.cn; dongjh8@mail2.sysu.edu.cn; xiexiaoh6@mail.sysu.edu.cn; stsljh@mail.sysu.edu.cn).}
	}
	
	\markboth{IEEE Transactions on Information Forensics and Security}%
	{Chen \MakeLowercase{\textit{et al.}}: Releasing Inequality Phenomenon in $\ell_{\infty}$-norm Adversarial Training via Input Gradient Distillation}
	
	
	\maketitle
	
	\begin{abstract}
		
		Adversarial training (AT) is considered the most effective defense against adversarial attacks. However, a recent study revealed that \(\ell_{\infty}\)-norm adversarial training (\(\ell_{\infty}\)-AT) will also induce unevenly distributed input gradients, which is called the inequality phenomenon. This phenomenon makes the \(\ell_{\infty}\)-norm adversarially trained model more vulnerable than the standard-trained model when high-attribution or randomly selected pixels are perturbed, enabling robust and practical black-box attacks against \(\ell_{\infty}\)-adversarially trained models. In this paper, we propose a simple yet effective method called Input Gradient Distillation (IGD) to release the inequality phenomenon in $\ell_{\infty}$-AT. IGD distills the standard-trained teacher model's equal decision pattern into the $\ell_{\infty}$-adversarially trained student model by aligning input gradients of the student model and the standard-trained model with the Cosine Similarity. Experiments show that IGD can mitigate the inequality phenomenon and its threats while preserving adversarial robustness. Compared to vanilla $\ell_{\infty}$-AT, IGD reduces error rates against inductive noise, inductive occlusion, random noise, and noisy images in ImageNet-C by up to 60\%, 16\%, 50\%, and 21\%, respectively. Other than empirical experiments, we also conduct a theoretical analysis to explain why releasing the inequality phenomenon can improve such robustness and discuss why the severity of the inequality phenomenon varies according to the dataset's image resolution. Our code is available at \url{https://github.com/fhdnskfbeuv/Inuput-Gradient-Distillation}
	\end{abstract}
	

	\section{Introduction}
	\label{sec:Intro} 
	\IEEEPARstart{I}{n} 2013,~\citet{firstadv} discovered the adversarial example, which can fool Deep Neural Networks (DNNs) by adding an imperceptible perturbation to the clean example. Since then, many adversarial defenses ~\cite{firstadv, PGD, mart, detect1, detect2} have been proposed to improve DNN's robustness against adversarial examples, among which adversarial training is considered the most effective. Besides improving DNN's adversarial robustness,~\citet{ATgoodness2} found that the \(\ell_{\infty}\)-norm adversarial training (\(\ell_{\infty}\)-AT) tends to produce sparse Integrated Gradients-based~\cite{IG} attribution map. They claimed that such sparseness means producing a concise explanation where only a few pixels have high attribution value and dominate DNN's output. Overall, suppressing non-robust or non-significant features was considered a benefit that $\ell_{\infty}$-AT brings.
	
	However, a recent study demonstrated that such suppression will result in unrealized threats. \citet{duaninequality} called such suppression the inequality phenomenon, and such a phenomenon makes the model less reliable. To be specific, they pointed out that $\ell_{\infty}$-adversarially trained models tend to produce attribution maps with higher Gini values~\cite{gini} than the standard-trained model's and are more vulnerable than the standard-trained model when a few pixels with high attribution values are perturbed by i.i.d. random noise or are occluded. Worse still, this vulnerability even exists when the perturbed pixels are randomly chosen, which allows the adversary to conduct a robust and practical black-box attack to the $\ell_{\infty}$-adversarially trained model without any information from the target model. While adversarial robustness is crucial for security, sacrificing robustness against practical and common perturbations, like i.i.d. random noise and occlusion, is not wise. Thus, our goals are to improve such practical robustness by releasing the inequality phenomenon and to preserve the $\ell_{\infty}$ adversarial robustness gained by $\ell_{\infty}$-AT.
	
	In this paper, we proposed a method called Input Gradient Distillation (IGD) to achieve our goals. During $\ell_{\infty}$-AT, IGD uses a standard-trained teacher model to generate equal input gradients as guidance and align the $\ell_{\infty}$-adversarially trained student model's input gradients with this guidance by Cosine Similarity, which allows IGD to distill the standard-trained model's equal decision pattern into the $\ell_{\infty}$-adversarially trained student model.
	
	Experimental results demonstrate that, on ImageNet-100, IGD can effectively release the inequality phenomenon in $\ell_{\infty}$-AT and improve the $\ell_{\infty}$-adversarially trained model's robustness against attacks devised by~\citet{duaninequality} while preserving its $\ell_{\infty}$ adversarial robustness. To be precise, concerning ImageNet-100, IGD reduces the error rate of the $\ell_{\infty}$-adversarially trained model from around 70\% to 10\% against inductive noise and from around 40\% to 24\% when confronted with inductive occlusion, and the IGD-trained model demonstrates good generalization to various occlusion colors, in contrast to CutOut ~\cite{cutout} used by~\citet{duaninequality}. We also test our method on noisy images of ImageNet-C ~\cite{imagenetC} and random noise~\cite{duaninequality}. Results show that, compared to the baseline, the model trained with IGD demonstrates better robustness against noisy images of ImageNet-C and random noise, with error rates drop of up to 21\% and 50\%, respectively.
	
	Besides releasing the inequality phenomenon, IGD also serves as a tool to analyze the inequality phenomenon because IGD can control variables like the Gini value and $\left\|\frac{\partial f^{y}(x)}{\partial x}\right\|_{1}$ which are properties of the attribution map. With the help of IGD, we conduct a theoretical analysis to explain why releasing the inequality phenomenon can improve such robustness. Furthermore, we find that the severity of the inequality phenomenon may vary according to the dataset's resolution (i.e., the number of pixels in an image) and explain why this phenomenon exists.
	
	The main contributions of our paper are as follows:
	\begin{itemize}
		\item We propose Input Gradient Distillation (IGD) to mitigate the inequality phenomenon and its threats in $\ell_{\infty}$-AT. Compared to the baseline, the IGD-trained model has a more even attribution map and better robustness against i.i.d. noise and occlusion while maintaining $\ell_{\infty}$ adversarial robustness gained by $\ell_{\infty}$-AT.
		
		\item We theoretically analyze the relationship between the Gini value~\cite{gini} of the input gradients and the model's robustness against noise and occlusion. We claim that the equality of the input gradients promotes the model's robustness by suppressing the deviation of the class score.
		
		\item We discover that the inequality phenomenon is more severe on high-resolution datasets than on low-resolution datasets. Based on empirical experiments and our theoretical analysis, we find that this property is common and explain why such a property exists.
	\end{itemize}

	\section{Related Work}
	\label{sec:related}
	
	\subsection{Adversarial Training for Security}
	
	Adversarial examples are inputs that can force the DNNs to generate incorrect predictions while remaining imperceptible to humans. Being the most effective adversarial defense against adversarial examples, adversarial training has received considerable attention from the information forensics and security community~\cite{at1, at2, at3, at4, at5, at6}. More recently, adversarial trained models are increasingly being integrated into Large Vision-Language Models (LVLM)~\cite{lvlm1, lvlm2, lvlm3} that serve as general-purpose systems. This trend not only expands the application scope of the adversarially trained models but also makes the robustness of the adversarially trained models more critical for security issues. In the rest of the paper, we focus on AT in image classification tasks because this task is the most common test-bed for studying adversarial robustness~\cite{Croce} and because new tasks concerning LVLM's robustness are still far from being well-defined~\cite{lvlmIll}.
	
	\subsection{Improving Robustness by Utilizing the Input Gradients}\label{sec:relatedGradient}
	
	Utilizing the input gradients for adversarial defenses has been proven to be effective. Previous works ~\cite{gradientRobustness1, gradientRobustness2} have attempted to improve DNN's adversarial robustness by simply penalizing the norm of input gradients. Other works, such as ~\cite{align1, align2}, aimed to align the input gradients of the model with some external guidance in order to obtain or enhance adversarial robustness. \citet{align3} aligned the input gradients between the benign input and the adversarial input to alleviate the catastrophic overfitting of the fast adversarial training. These studies demonstrate that incorporating input gradients into the loss function can effectively enhance the model's performance. However, these works can not address the inequality phenomenon we discussed off-the-shelf because they either lack competitive adversarial robustness\cite{gradientRobustness1, gradientRobustness2} or align the model's input gradients to the $\ell_{\infty}$ adversarial trained model's~\cite{align1, align2, align3} with inappropriate metric~\cite{align2}, which is opposite to our goal.
	
	\section{{Background}}
	\label{sec:background}
	
	\subsection{$\ell_{\infty}$-AT}\label{sec:introAT}
	
	The main idea of adversarial training is adding adversarial examples in the training phase. \citet{PGD} improved DNN's adversarial robustness by solving a min-max optimization problem:
	\begin{alignat}{2}
		\min_{\theta}\max_{\delta} \quad &\mathcal{L}(f_{\theta }(x+\delta ), y)\\
		\nonumber\mathrm{s.t.} \quad &\left \| \delta \right \|_{p} \le \epsilon\,,
	\end{alignat}
	where $\delta$ is the adversarial perturbation, $x$ is the benign example, $y$ is the label, $f_{\theta}(\cdot)$ is the DNN with weight $\theta$, $\mathcal{L}(\cdot, \cdot)$ is the loss function, and the inner maximum problem is solved by the Projected Gradient Descent (PGD)~\cite{PGD}. If we define $p=\infty$, then this method is called $\ell_{\infty}$-AT which is the most widely studied adversarial training method according to the number of reported results in RobustBench\cite{robustbench}. 
	
	\subsection{Inequality phenomenon in $\ell_{\infty}$-AT}
	\label{sec:introInequal}
	\citet{ATgoodness2} found that, compared to the standard training, $\ell_{\infty}$-AT will make DNN produce a more unequal input attribution map\footnote{We use the input gradients to represent the input attribution map by default. We discuss other attribution methods in \cref{sec:otherAM}.} ($A^{f}(x)$ for brevity where $f$ stands for DNN and $x$ stands for the input), where a few pixels have much higher attribution values than others. \citet{duaninequality} named this property inequality phenomenon in $\ell_{\infty}$-AT. To quantitatively measure the inequality of the input attribution map, \citet{duaninequality} proposed to use the Gini value~\cite{gini} 
	\begin{equation}
		\label{eqn:GiniEq}
		\Gini(\Phi )=\frac{1}{n}*(n+1 -2*\frac{\sum_{i=1}^{n} (n+1-i)*\phi_{i} }{\sum _{i=1}^{n}\phi _{i}} )\,, 
	\end{equation}
	where $\Phi = \left\{\phi_{i}, i=1...n \mid 0 \le \phi_{i} \le \phi_{i+1} \right\}$. The lower the Gini value is, the more even the $\Phi$'s distribution is. Specifically, when measuring the inequality of $A^{f}(x)$, \citet{duaninequality} first take the absolute value of $A^{f}(x)$ and sort $|A^{f}(x)|$ in ascending order. Then, treating the sorted $|A^{f}(x)|$ as $\Phi$, one can calculate the Gini value of $A^{f}(x)$ using \cref{eqn:GiniEq}.
	
	To better understand and measure the inequality phenomenon,~\citet{duaninequality} defined two types of inequality: global inequality and regional inequality. These two types of inequality are both measured by the Gini value. They calculate the Gini value of $A^{f}(x)$ ($\Gini(A^{f}(x)$) for short) to measure global inequality. Global inequality can reflect whether the model makes decision based on a few pixels with high attribution values. For regional inequality, they divide $A^{f}(x)$ into blocks with size $l \times l$, sum up values within the block to get a downsampled attribution map ($A_{r}^{f}(x)$ for short where $r$ stands for the region) and calculate the Gini value of $A_{r}^{f}(x)$ ($\Gini(A_{r}^{f}(x))$ for short). Unlike global inequality, regional inequality introduces locality into measuring inequality and can reflect whether important pixels tend to cluster in a few regions.
	
	To reveal threats brought by the inequality phenomenon,~\citet{duaninequality} devised two attack algorithms called Inductive Noise Attack and Inductive Occlusion Attack to attack the $\ell_{\infty}$-adversarially trained model.
	
	\subsubsection{\textbf{Inductive Noise Attack}}
	
	Inductive Noise Attack (INA) perturbs the input with Gaussian noise $\delta \in \mathcal{N} (0,1)$ in an order determined by pixels' attribution values $a_{i} \in A^{f}(x)$. Formally, we have
	\begin{align}
		\label{eqn:ina1}
		x' = x + M * \delta, \; \text{where} \; M_{i}=\begin{cases}
			1,\; &a_{i} > threshold;\\
			0,\; &a_{i} \le threshold.
		\end{cases}\,.
	\end{align}
	Besides the additive noise introduced above,~\citet{duaninequality} also devised a noise attack where the original pixel is replaced by noise:
	\begin{align}
		\label{eqn:ina2}
		x' = \bar{M} * x + M * \delta, \; \text{where}\; \bar{M} = 1 - M\,.
	\end{align}
	For conciseness, we refer to the former INA as ``\emph{INA1}'' and to the latter INA as ``\emph{INA2}''. The numbers of perturbed pixels are identical between different evaluated models for fair comparisons. Thus, if we want to perturb $k$ pixels, the $threshold$ is determined by the $k$th largest attribution value.
	
	\subsubsection{\textbf{Inductive Occlusion Attack}}
	Inductive Occlusion Attack (IOA) progressively occludes regions with high attribution values. In each iteration, IOA selects the $n$ biggest pixels as regions' central points, occludes these regions on the clean input with a $(2r+1) \times (2r+1)$ pure color patch, and feeds the perturbed input into the model.~\citet{duaninequality} set $N$ and $R$ to limit $n$ and $r$ and chose black, gray, or white colors to occlude the image. IOA increases $r$ if the model is not fooled in the current iteration, increases $n$ and sets $r$ to $1$ if $r$ exceeds $R$, and stops if the model is fooled or both $n$ and $r$ exceed their limitations. We refer to IOA with black, gray, and white patches as IOA-B, IOA-G, and IOA-W, respectively.
	
	Besides two inductive methods mentioned above,~\citet{duaninequality} also used random noise (RN) to attack DNN, which randomly selects $k$ pixels and perturbs them with Gaussian noise $\delta \in \mathcal{N} (0,1)$.
	
	For conciseness, we refer to the model's robustness against the attacks mentioned above as \emph{inequality-based robustness} in the rest of the paper.

	\section{Methodology}
	\label{sec:Method}
	In this section, we first introduce our motivation behind using Cosine Similarity to align the input gradients between the standard-trained model and the $\ell_{\infty}$-adversarially trained model. Then, we introduce the framework of IGD.
	
	\subsection{Motivation} \label{sec:motivation}
	Recall our goals mentioned in Section~\ref{sec:Intro} that we intend to release the inequality phenomenon in $\ell_{\infty}$-AT and to maintain its adversarial robustness. One reasonable approach to achieve these goals is aligning the input gradients to guidance with a more even distribution. Recall \cref{sec:relatedGradient} that aligning input gradients has been proven effective in distilling the knowledge of the teacher model into the student model. However, the guidance and alignment metric they used are inappropriate for addressing the inequality phenomenon. Thus, choosing appropriate guidance and a suitable metric for alignment is the key to designing our method.
	
	To generate guidance with a more even distribution, the standard-trained model can be a good choice because \citet{duaninequality} found that the distribution of the standard-trained model's input gradients is more even than the $\ell_{\infty}$-adversarially trained model's. Moreover, we can obtain the standard-trained model with relatively low costs. Thus, we choose the standard-trained model to generate the guidance.
	
	After determining the guidance for the input gradient, we should choose a proper metric to align them. Such a metric should have two properties. First, the metric should align the Gini value between them, and second, optimizing the metric should not degrade the student model's adversarial robustness significantly. Cosine Similarity can be an appropriate metric satisfying these two properties.
	
	First, the Gini value and the Cosine Similarity are both determined by the vector's direction, i.e., the components' relative value, rather than its norm. If we align two vectors' direction and have $\cos(\vec{v}_{1}, \vec{v}_{2}) = 1$, which indicates that there exists $k \in \mathbb{R}$ such that $\vec{v}_{1} = k * \vec{v}_{2}$, then we have $\Gini(\vec{v}_{1})=\Gini(\vec{v}_{2})$. This property means that if we want to match two vectors' Gini values, we only need to align their directions, and aligning their norm does not contribute to matching their Gini values.
	
	Second, using Cosine Similarity to align two vectors will not alter their norm significantly because Cosine Similarity fixes $\left\|\vec{v}\right\|_{2} = 1$ with $L_{2}$ normalization and will not penalize the norm of vectors. Such a property is crucial for adversarial robustness because previous works~\cite{gradientRobustness1, gradientRobustness2} have claimed that adversarial robustness has a negative correlation with the norm of the input gradients. Since the norm of the standard-trained model's input gradients is much larger than that of the $\ell_{\infty}$-adversarially trained model (\textbf{see the last column of \cref{tab:advGinitable}}), aligning their norm may degrade the adversarial robustness. Furthermore, this property can also help us control variables like $\left\|\frac{\partial f^{y}(x)}{\partial x}\right\|_{1}$, and only changes the distribution of the component in the input gradient, which is of great importance to our analysis of the inequality phenomenon in Section~\ref{sec:explain} and Section~\ref{sec:assmption}.
	
	To sum up, we use a standard-trained model as the teacher model to generate guidance for the input gradients and distill its equal decision pattern into the $\ell_{\infty}$-adversarially trained student model by leveraging Cosine Similarity to align their input gradients.

	\begin{figure*}[!t]
		\centering
		\includegraphics[width=\linewidth]{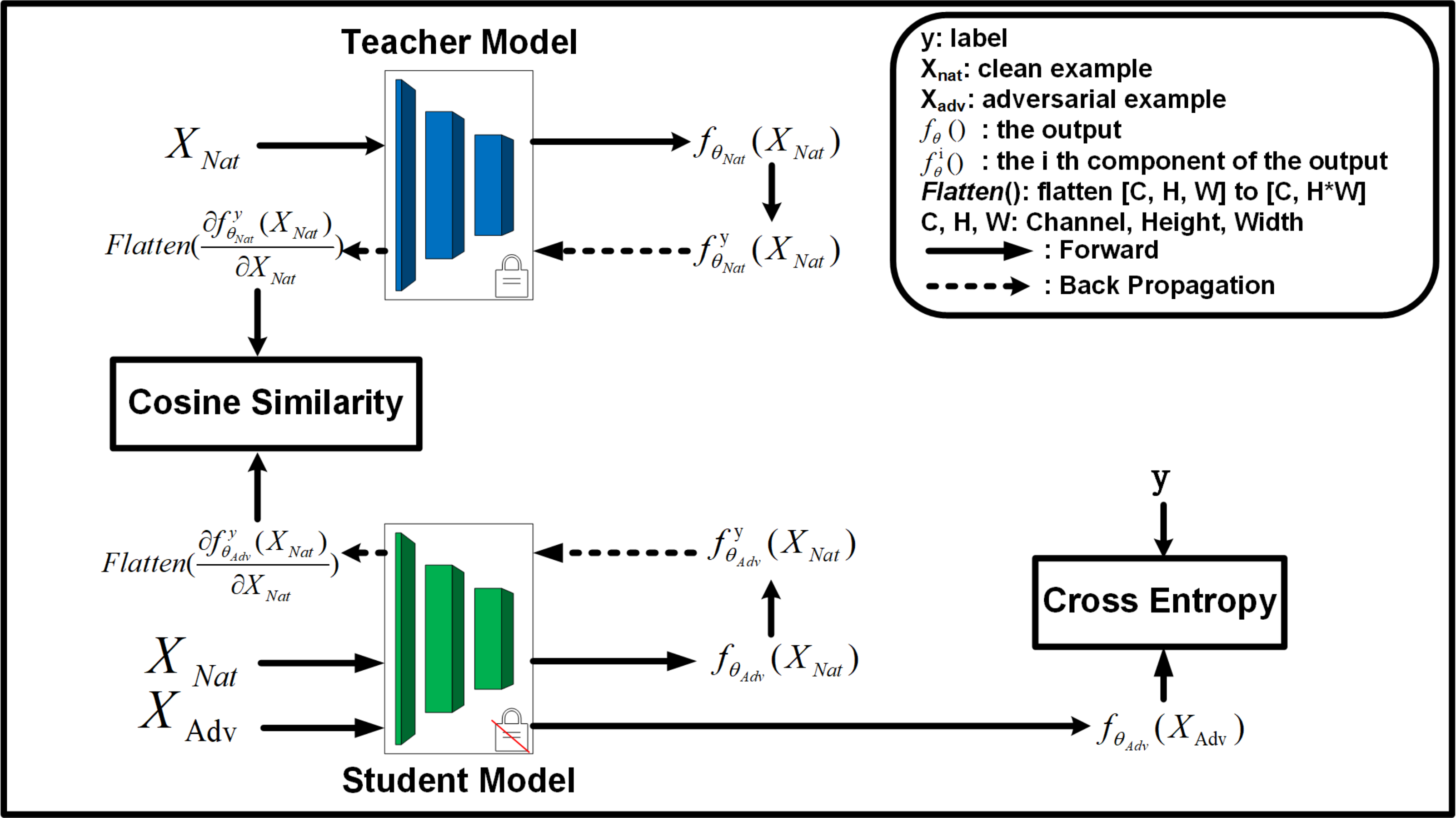}
		\caption{The framework of IGD.}
		\label{fig:framework}
	\end{figure*}
	
	\subsection{Input Gradient Distillation}
	The framework of the Input Gradient Distillation (IGD) is illustrated in Figure~\ref{fig:framework}. The key component of IGD is the alignment of input gradients between the standard-trained teacher model and the $\ell_{\infty}$-adversarially trained student model. In each iteration, the clean example is fed into the student model and the teacher model, and their input gradients are aligned by minimizing
	\begin{align}
		\mathcal{L}_{align} &= \cos(\Flatten(\frac{\partial f_{\theta_{Adv}}^{y}(x)}{\partial x}),  \Flatten(\frac{\partial f_{\theta_{Std}}^{y}(x)}{\partial x}))\,,\label{eq:alignLoss}
	\end{align}
	where $x$ is the clean example, $y$ is the label, $f_{\theta}^{i}(\cdot)$ is the $i$th component of the output logit, $\cos(\cdot, \cdot)$ is the Cosine Similarity, and $\Flatten(\cdot)$ is a reshaping function that reshapes the input with shape $[Channel, Height, Width]$ to $[Channel, Height*Width]$. Note that $\cos(\cdot, \cdot)$ will align the spatial distribution of the input gradients because $\Flatten(\cdot)$ flattens the last two channels which represent the input's height and width. The other component of IGD is a vanilla PGDAT~\cite{PGD} whose loss function is
	\begin{align}
		\mathcal{L}_{ce} = \text{CE}(f_{\theta_{Adv}}(x'), y)\,,
	\end{align}
	where $x'$ is the adversarial example generated by PGD, and $\text{CE}(\cdot, \cdot)$ is the Cross-Entropy loss. We choose PGDAT as it does not introduce additional regularizers that may couple with ours, which can be beneficial for our analysis. To sum up, the overall loss is
	\begin{equation}
		\mathcal{L} = \mathcal{L}_{ce} - \lambda * \mathcal{L}_{align}\,,
	\end{equation}
	where $\lambda$ is a hyper-parameter to balance $\mathcal{L}_{ce}$ and $\mathcal{L}_{align}$.

	\section{Experiments}
	\label{sec:Experiment}
	
	In this section, we first describe the settings of our experiments. Then, we evaluate IGD's performance in alleviating the inequality phenomenon and its threats by evaluating models' inequality-based robustness and visualizing their attribution maps.
	
	\subsection{Experimental Setting}
	\label{sec:experSetup}
	
	\subsubsection{\textbf{Datasets}} \label{sec:dataset}
	
	We select CIFAR100 and ImageNet-100 for our experiments. The CIFAR100 contains 60,000 RGB images of size \(32 \times 32\) pixels. CIFAR-100 has 50,000 training and 10,000 test images, spanning 100 classes. The ImageNet-100 dataset is a 100-class subset of a large-scale visual recognition benchmark, ImageNet~\cite{imagenet}. The ImageNet-100 has 12k training images and 5k validation images, having a higher resolution than the CIFAR family. We resize and crop all ImageNet-100 images to a fixed resolution of \(224 \times 224\) pixels.
	
	\subsubsection{\textbf{Models}} \label{sec:model}
	We adopt ResNet18 \cite{resnet} as the backbone model for our experiments. ResNet-18 is a widely used convolutional neural network (CNN) with 18 layers, designed based on residual learning to facilitate training deeper networks. ResNet18 follows a hierarchical structure with four residual blocks, each containing two convolutional layers, and utilizes global average pooling followed by a fully connected layer for classification. The model is computationally efficient while maintaining strong performance, making it a suitable choice for evaluating robustness across different datasets and training paradigms.
	
	\subsubsection{\textbf{Attribution Methods}} \label{sec:am}
	
	We use Input X Gradients~\cite{inputx}, Integrated Gradients~\cite{IG}, Shapley Value~\cite{gradShap}, SmoothGrad~\cite{smoothGrad}, and Saliency Map~\cite{saliency} provided by \citet{captum} to generate attribution maps. We mainly use Saliency Map~\cite{saliency}, which treats the input gradients as the attribution map, and will note if other attribution methods are used.
	
	\subsubsection{\textbf{Baselines}}
	The inequality phenomenon and its threats are newly discovered by~\citet{duaninequality}, and ours is the second work addressing this issue. Thus, to the best of our knowledge, there is no related defense proposed by any previous work other than combining CutOut~\cite{cutout} with PGDAT~\cite{PGD}, which is proposed by~\citet{duaninequality}. The CutOut size is $64\times64$ for ImageNet100 and $16\times16$ for CIFAR. To conclude, the baselines we mainly discuss are PGDAT~\cite{PGD} and PGDAT+CutOut~\cite{duaninequality}.

	\subsubsection{\textbf{Training setup}} For fairness, all models are trained for 150 Epochs by Stochastic Gradient Descent (SGD) with momentum 0.9 and weight decay 5e-4. The batch size is 128. The initial learning rate is 0.1 and is divided by 10 on the plateau of training loss. Unless other specified, when crafting adversarial examples, $\epsilon$ is set to $8/255$, the step size is set to $2/255$, and the iteration step is set to $10$.

	\subsubsection{\textbf{Evaluation setup}}To evaluate the adversarial robustness fairly, we use AutoAttack~\cite{autoattack} with $\epsilon=8/255$. We use error rate to evaluate models' robustness against attacks introduced in Section~\ref{sec:introInequal} and noisy images of the subset of ImageNet-C~\cite{imagenetC} where we select the same classes as ImageNet-100's. The error rate is the proportion of misclassified examples among examples correctly classified by all models being compared. We follow the settings of the attack algorithm in ~\cite{duaninequality} and will clarify attack parameters if necessary. When using IOA, we set N=10 and R=4 for CIFAR100 and R=20 for ImageNet-100. We set $r=16$ for ImageNet-100 and $r=4$ for CIFAR100 when measuring regional inequality.
	
	\subsubsection{\textbf{Visualization setup}} We first take the absolute value of the attribution map and calculate the mean value of the channel dimension to acquire a mono-channel attribution map. Then, we clip pixels of the attribution map to within $\pm3\sigma$ and rescale the pixels' value to $[0,1]$. Finally, we map the mono-channel attribution map to a 3-channel RGB image, \emph{where warm colors represent high values and cold colors represent low values}.
	\subsection{Releasing the inequality phenomenon}\label{sec:releaseInequality}
	\begin{table*}[!t]
		\centering
		\caption{Standard accuracy, adversarial accuracy, global Gini value, regional Gini value, and $\left\|\frac{\partial f^{y}(x)}{\partial x}\right\|_{1}$ of ResNet18 across different datasets and methods.}
		\begin{tabular}{@{}ccccccc@{}}
			\toprule
			Dataset  & Method           & Std. Acc. $\uparrow$ & Adv. Acc. $\uparrow$ & $\Gini(A^{f}(x))$            & $\Gini(A_{r}^{f}(x))$            & $\left\|\frac{\partial f^{y}(x)}{\partial x}\right\|_{1}$              \\ \midrule
			\multirow{7}{*}{CIFAR100}     & Standard         & 77.56\%  & 0.00\%   & 0.530 (80\%)  & 0.342 (65\%)  & 2,756.37 (100\%) \\
			& PGDAT\cite{PGD}           & 57.62\%  & 24.61\%  & 0.666 (100\%) & 0.527 (100\%) & 66.04 (2.40\%)  \\
			& PGDAT+CutOut\cite{duaninequality}    & 53.77\%  & 23.62\%  & 0.665 (100\%) & 0.524 (99\%)  & 57.07 (2.07\%)  \\
			& IGD ($\lambda$=1) & 58.04\%  & 24.83\%  & 0.663 (100\%) & 0.526 (100\%) & 65.28 (2.37\%)  \\
			& IGD ($\lambda$=2) & 57.79\%  & 24.33\%  & 0.638 (96\%)  & 0.507 (96\%)  & 61.66 (2.24\%)  \\
			& IGD ($\lambda$=3) & 56.39\%  & 23.45\%  & 0.617 (93\%)  & 0.489 (93\%)  & 57.36 (2.08\%)  \\
			& IGD ($\lambda$=4) & 57.02\%  & 22.43\%  & 0.603 (91\%)  & 0.475 (90\%)  & 62.96 (2.28\%)  \\ \midrule
			\multirow{7}{*}{ImageNet-100} & Standard         & 87.32\%  & 0.00\%      & 0.544 (58\%)  & 0.328 (58\%)  & 4,596.42 (100\%) \\
			& PGDAT\cite{PGD}           & 70.74\%  & 33.28\%  & 0.933 (100\%) & 0.565 (100\%) & 89.31 (1.94\%)  \\
			& PGDAT+CutOut\cite{duaninequality}    & 70.36\%  & 32.54\%  & 0.924 (99\%)  & 0.558 (99\%)  & 91.09 (1.98\%)  \\
			& IGD ($\lambda$=1) & 71.42\%  & 33.02\%  & 0.834 (89\%)  & 0.557 (99\%)  & 92.70 (2.02\%)  \\
			& IGD ($\lambda$=2) & 71.92\%  & 32.28\%  & 0.737 (79\%)  & 0.554 (98\%)  & 93.37 (2.03\%)  \\
			& IGD ($\lambda$=3) & 72.10\%  & 31.82\%  & 0.705 (76\%)  & 0.536 (95\%)  & 93.64 (2.04\%)  \\
			& IGD ($\lambda$=4) & 72.18\%  & 31.20\%  & 0.694 (74\%)  & 0.532 (94\%)  & 92.01 (2.00\%)  \\ \bottomrule
		\end{tabular}
		
		\label{tab:advGinitable}
	\end{table*}
	\begin{figure}[!t]
		\centering
		\subfloat[Saliency Map]{
			\centering
			\includegraphics[width=0.18\linewidth]{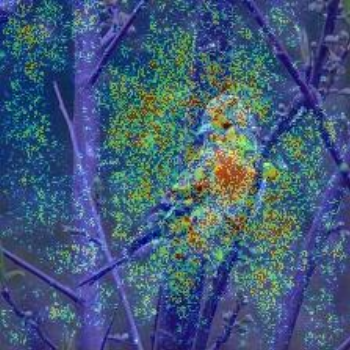}
			\includegraphics[width=0.18\linewidth]{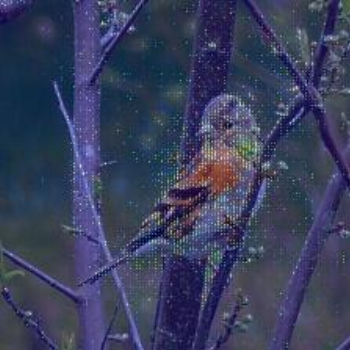}
			\includegraphics[width=0.18\linewidth]{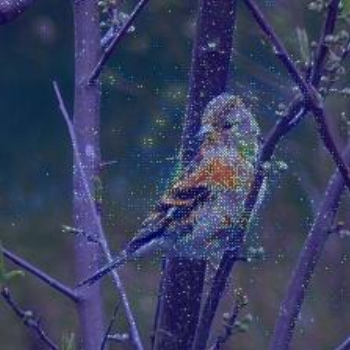}
			\includegraphics[width=0.18\linewidth]{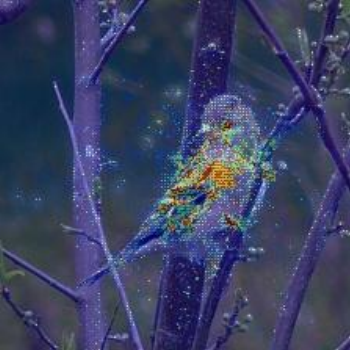}
			\includegraphics[width=0.18\linewidth]{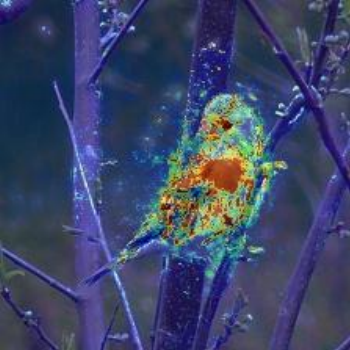}
		}\\
		\subfloat[Integrated Gradients]{
			\centering
			\includegraphics[width=0.18\linewidth]{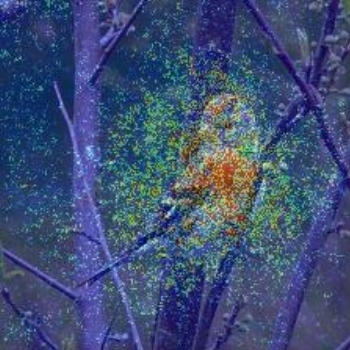}
			\includegraphics[width=0.18\linewidth]{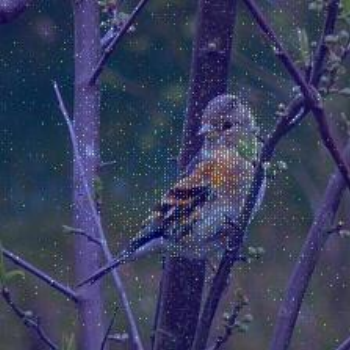}
			\includegraphics[width=0.18\linewidth]{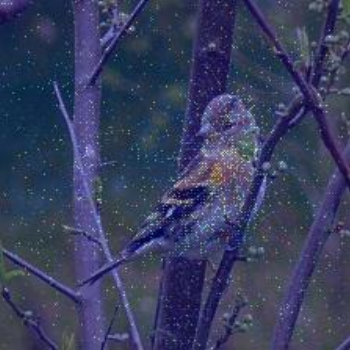}
			\includegraphics[width=0.18\linewidth]{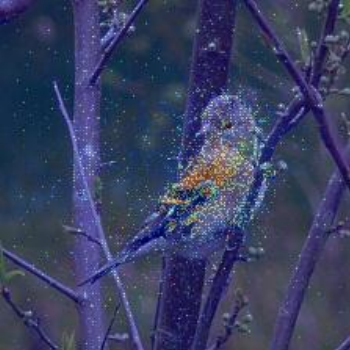}
			\includegraphics[width=0.18\linewidth]{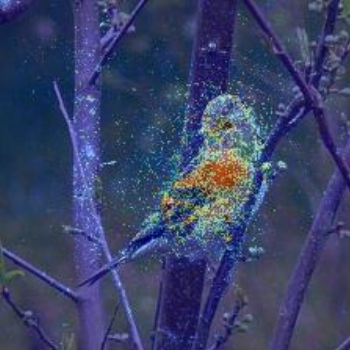}
		}\\
		\subfloat[Shapley Value]{
			\centering
			\includegraphics[width=0.18\linewidth]{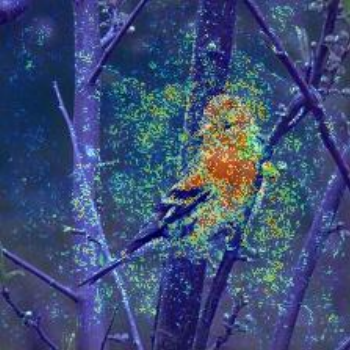}
			\includegraphics[width=0.18\linewidth]{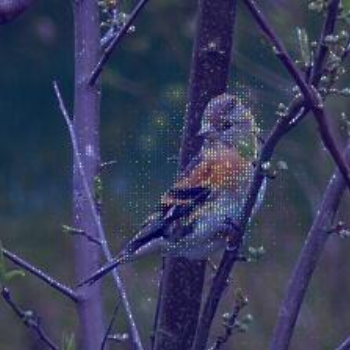}
			\includegraphics[width=0.18\linewidth]{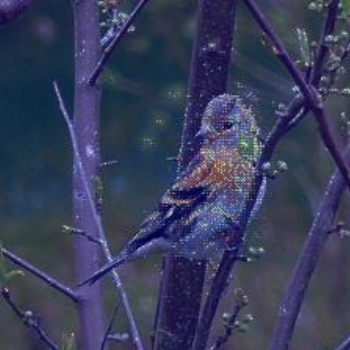}
			\includegraphics[width=0.18\linewidth]{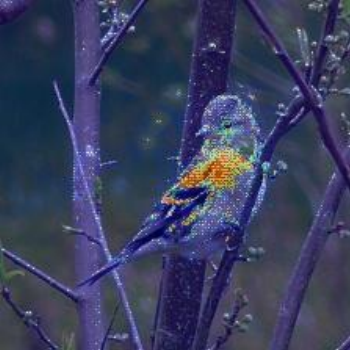}
			\includegraphics[width=0.18\linewidth]{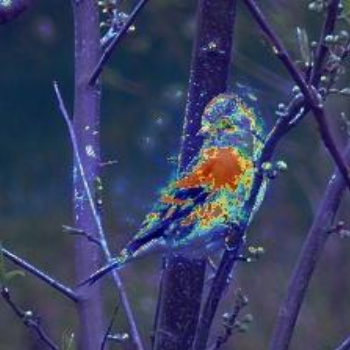}
		}\\
		\subfloat[Input X Gradients]{
			\centering
			\includegraphics[width=0.18\linewidth]{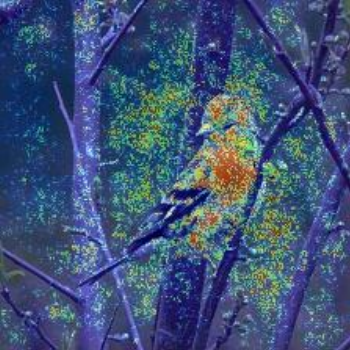}
			\includegraphics[width=0.18\linewidth]{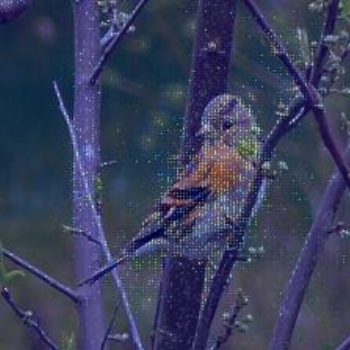}
			\includegraphics[width=0.18\linewidth]{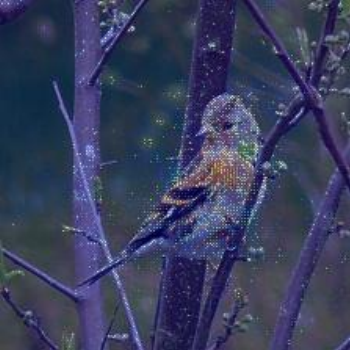}
			\includegraphics[width=0.18\linewidth]{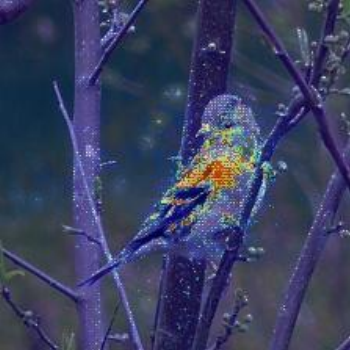}
			\includegraphics[width=0.18\linewidth]{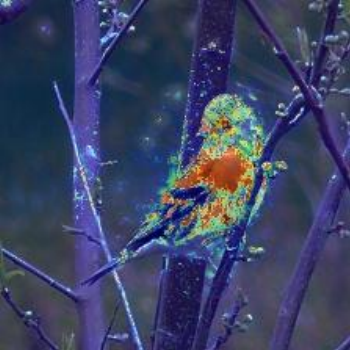}
		}\\
		\subfloat[SmoothGrad]{\label{visSmooth}
			\centering
			\includegraphics[width=0.18\linewidth]{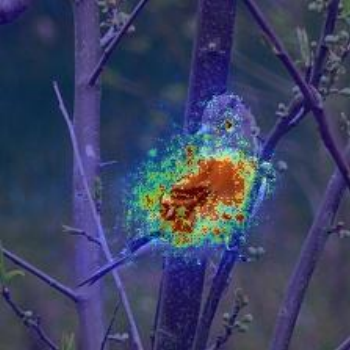}
			\includegraphics[width=0.18\linewidth]{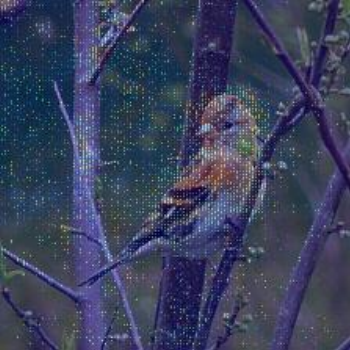}
			\includegraphics[width=0.18\linewidth]{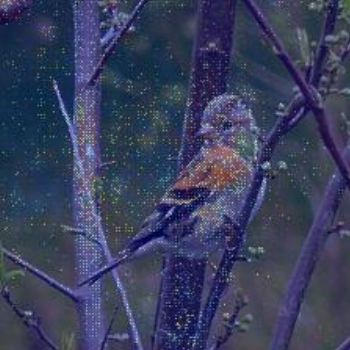}
			\includegraphics[width=0.18\linewidth]{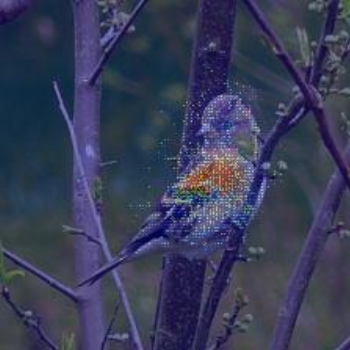}
			\includegraphics[width=0.18\linewidth]{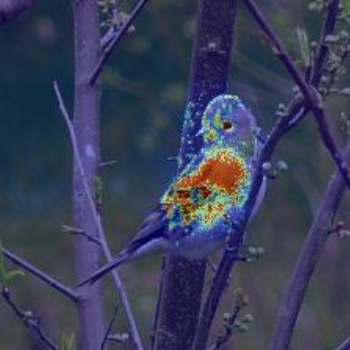}
		}
		\caption{Visualizations of attribution maps on ImageNet-100 across different attribution methods. From left to right: Standard training, PGDAT, PGDAT+CutOut, IGD ($\lambda=1$), IGD ($\lambda=4$). Best viewed in color.}
		\label{fig:imagenet100Visual}
	\end{figure}
	
	\begin{figure*}[!t]
		\begin{minipage}{\linewidth}
			\centering
			\includegraphics[width=0.7\linewidth]{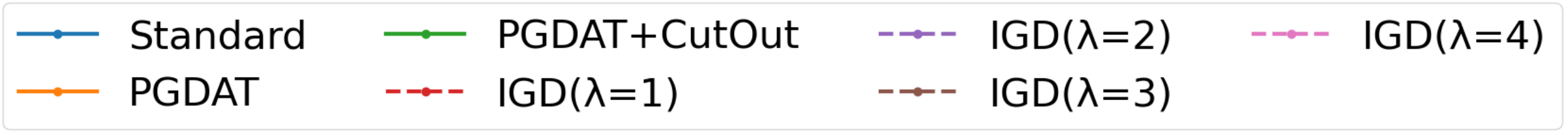}\\
			\subfloat[INA1 on ImageNet-100]{
				\label{imagenet100NoiseA}
				\includegraphics[width=0.32\linewidth]{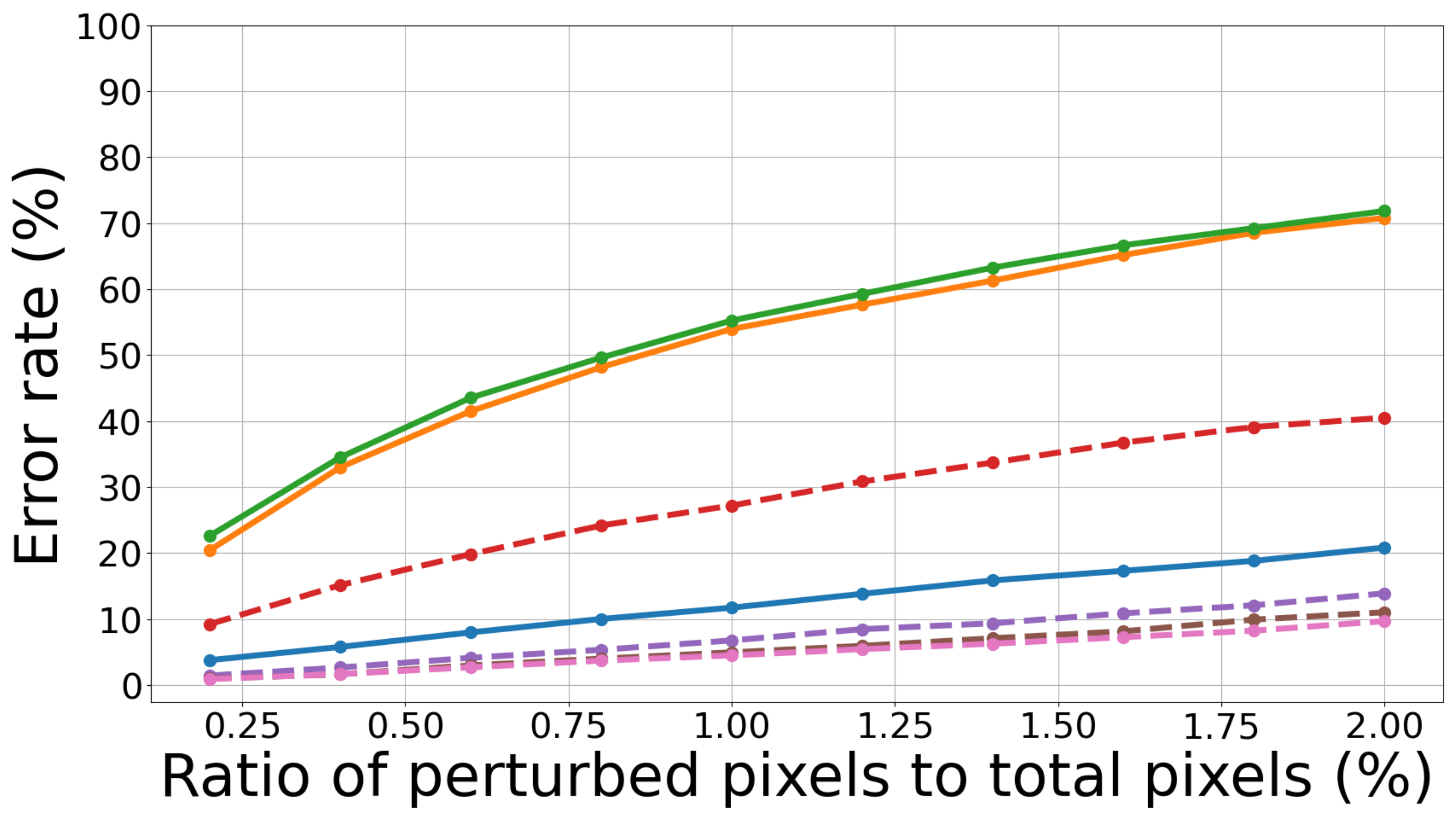}
				
			}
			\subfloat[INA2 on ImageNet-100]{
				\label{imagenet100NoiseB}
				\includegraphics[width=0.32\linewidth]{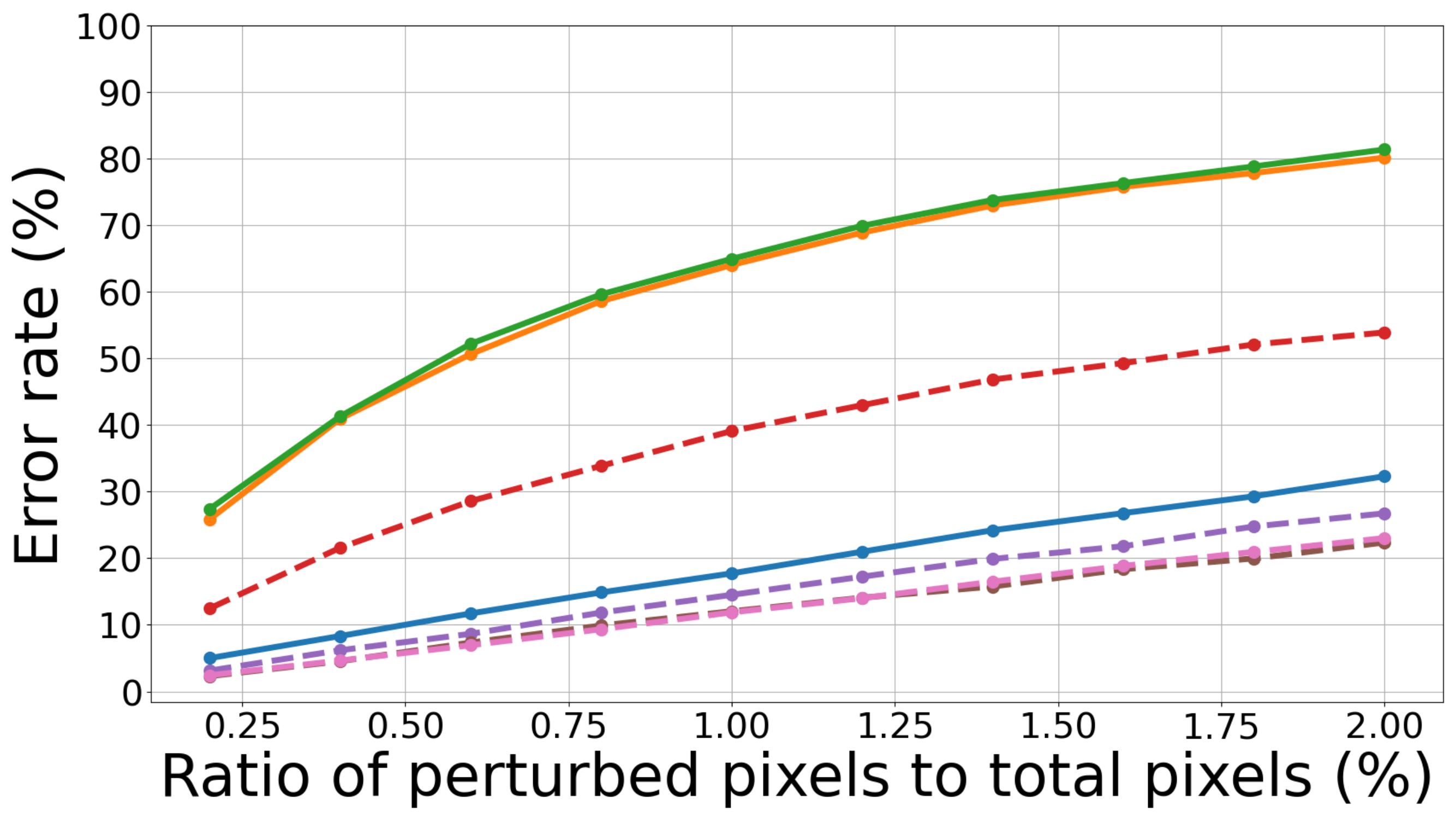}
				
			}
			\subfloat[RN on ImageNet-100]{
				\label{imagenet100NoiseC}
				\includegraphics[width=0.32\linewidth]{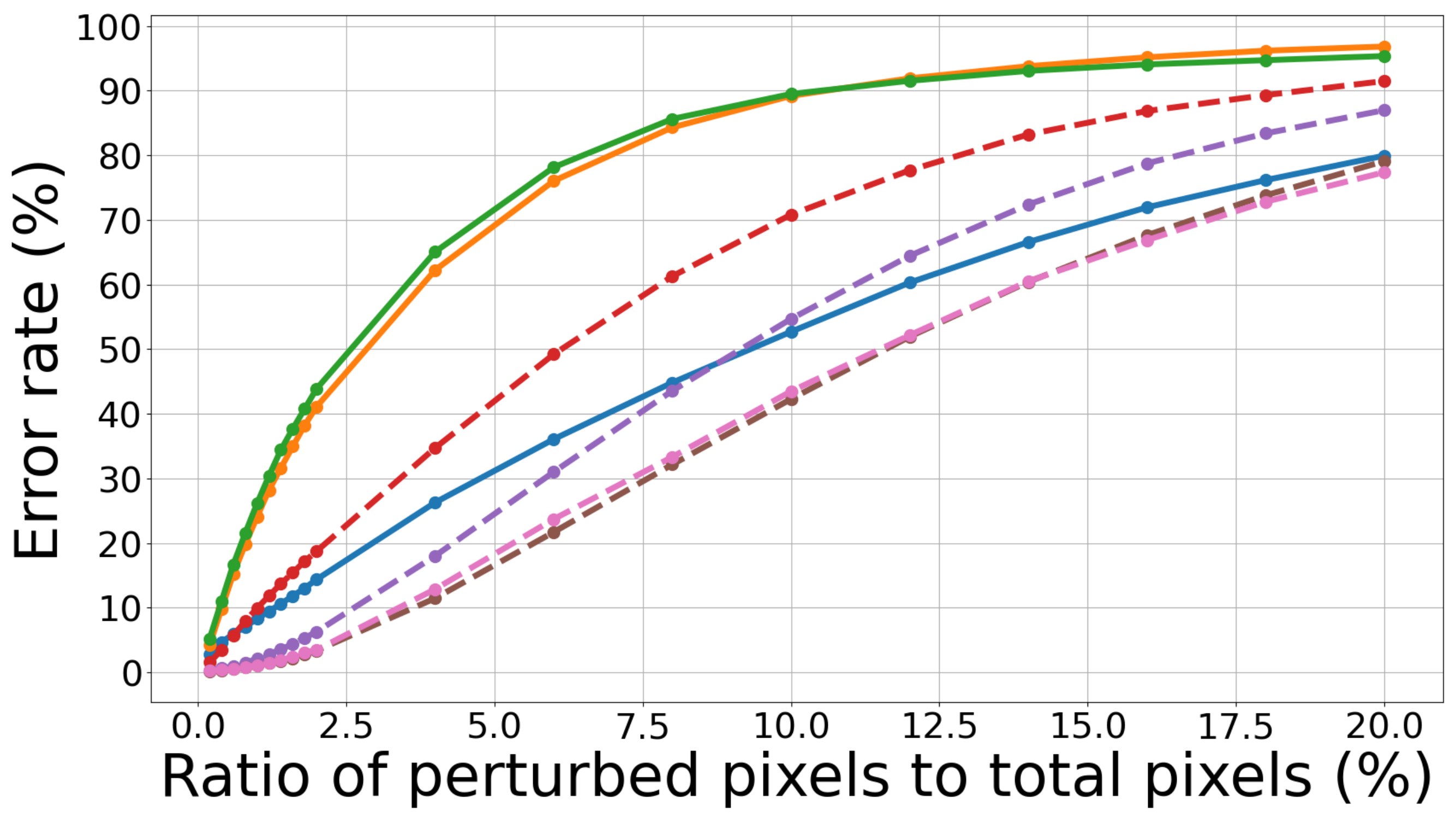}
				
			}
		\end{minipage}\\
		\begin{minipage}{\linewidth}
			\centering
			\subfloat[INA1 on CIFAR100]{
				\label{cifar100NoiseA}
				\includegraphics[width=0.32\linewidth]{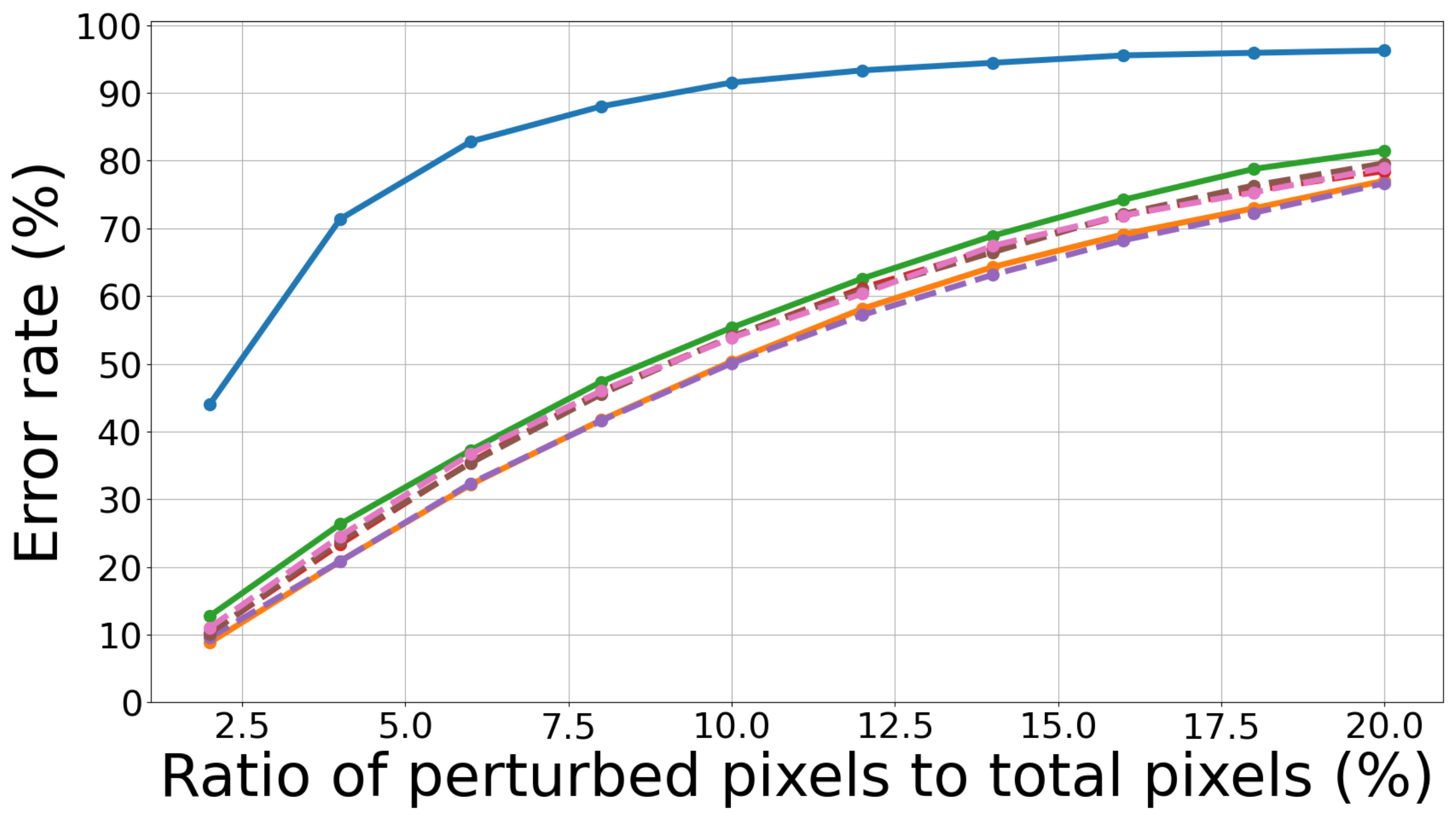}
				
			}
			\subfloat[INA2 on CIFAR100]{
				\label{cifar100NoiseB}
				\includegraphics[width=0.32\linewidth]{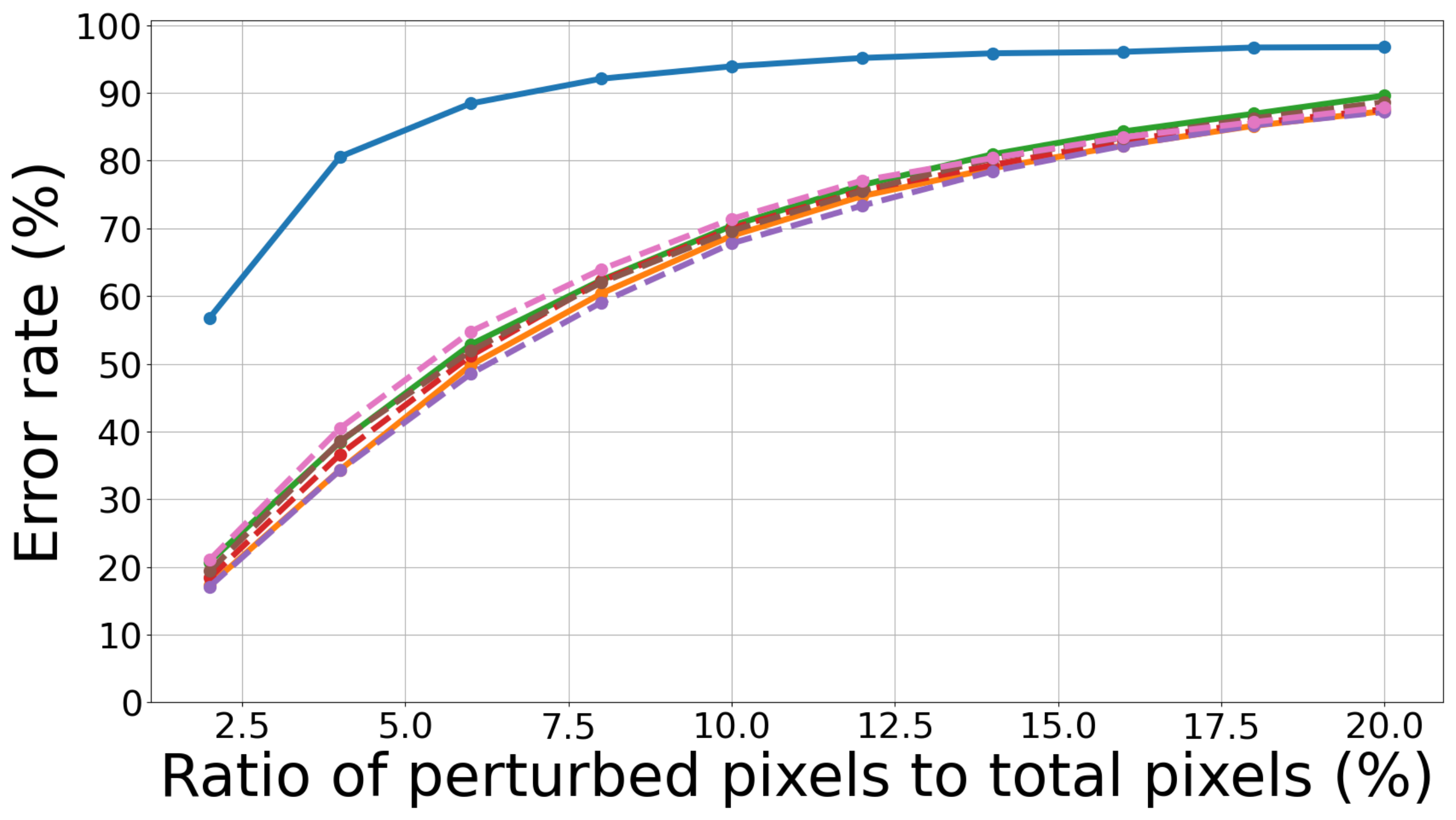}
				
			}
			\subfloat[RN on CIFAR100]{
				\label{cifar100NoiseC}
				\includegraphics[width=0.32\linewidth]{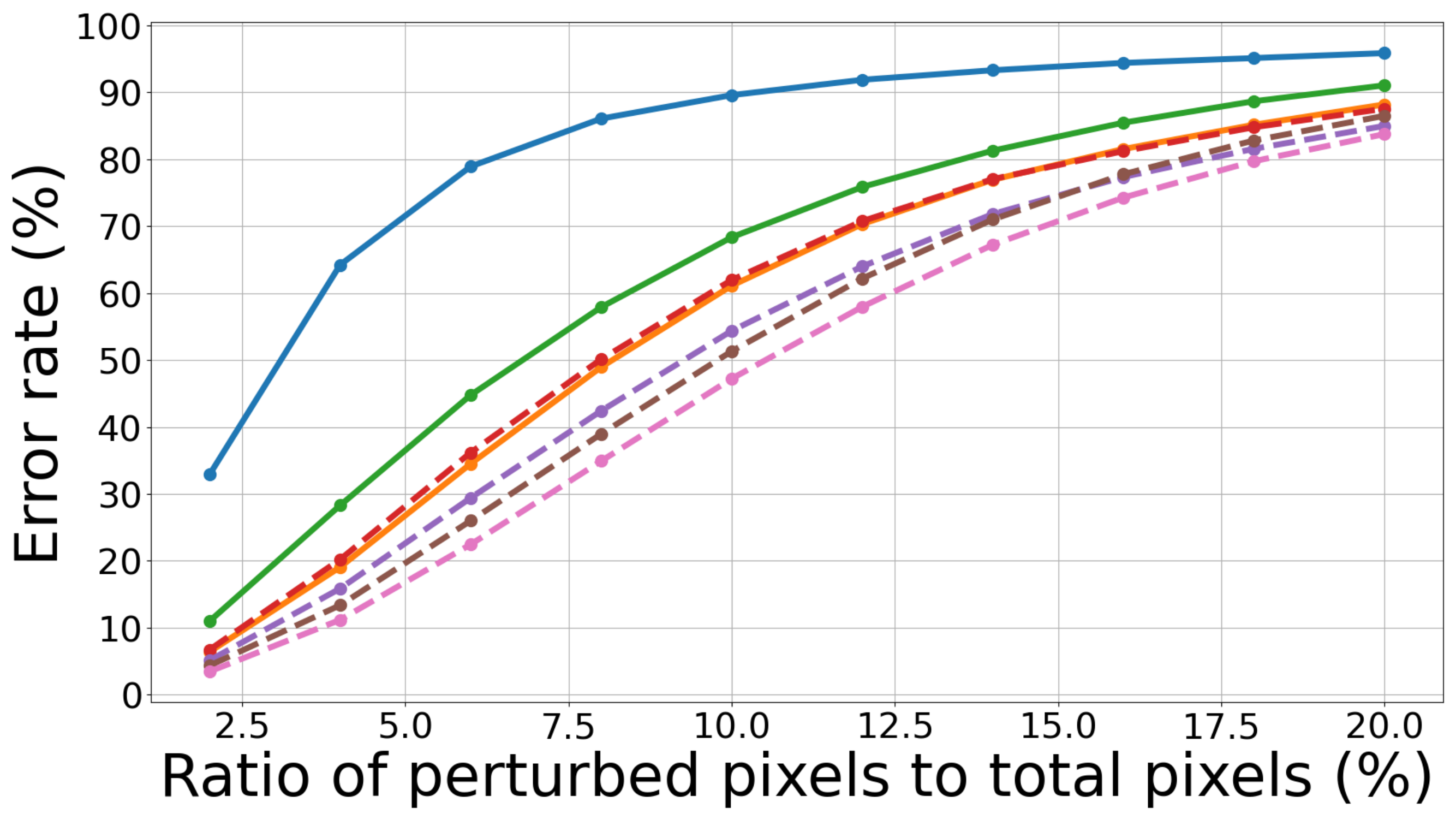}
				
			}
			\caption{Error rates $\downarrow$ of different methods against INA or RN on ImageNet-100 and CIFAR100. We can find that the threats of the inequality phenomenon are more obvious on high-resolution dataset ImageNet-100 than low-resolution dataset CIFAR100. Best viewed in color.}\label{fig:allNoise}
		\end{minipage}
	\end{figure*}
	In this section, we demonstrate the effectiveness of IGD in releasing the inequality phenomenon and comparisons to other methods. Results in Table~\ref{tab:advGinitable} show that IGD can effectively release the global inequality in $\ell_{\infty}$-AT and control the level by tuning coefficient $\lambda$. On ImageNet-100, IGD can decrease PGDAT's $\Gini(A^{f}(x))$ by up to 26\% and decrease PGDAT's $\Gini(A^{f}(x))$ by up to 9\% on CIFAR100, while CutOut can only slightly release the global inequality by 1\%. In Table~\ref{tab:advGinitable}, compared to the PGDAT-trained model, IGD does not change $\left\|\frac{\partial f^{y}(x)}{\partial x}\right\|_{1}$ a lot, which accords with our assumption that IGD aligns the direction and does not alter the norm of the input gradients significantly.
	
	Though IGD can release global inequality well, it can not release regional inequality effectively, suggesting that pixels with a high attribution value tend to cluster in a few regions. As for preserving the model's adversarial robustness, the adversarial accuracy only drops 1-2\% on ImageNet-100 and CIFAR100.

	To provide a more intuitive explanation of how IGD works, we visualize attribution maps on ImageNet-100 in Figure~\ref{fig:imagenet100Visual}. Pixels with warm colors on visualized attribution maps have high attribution values and dominate the prediction~\cite{duaninequality}. In Figure~\ref{fig:imagenet100Visual}, warm color pixels in attribution maps of PGDAT are much fewer than the IGDs', which indicates that the IGD-trained models have a relatively equal decision pattern and explains IGD-trained models' lower $\Gini(A^{f}(x))$. Regarding the PGDAT-trained model with CutOut, its attribution map does not exhibit any noticeable visual differences from the PGDAT-trained model, which can support our findings that CutOut can not release the global inequality. We can also find that warm color pixels on the background in IGDs' attribution maps are fewer than the standard's. The warm color pixels mainly lie on perceptually-aligned areas like bird wings. This outcome explains IGD's relatively high $\Gini(A_{r}^{f}(x))$, as important pixels tend to cluster in the object area. However, with larger $\lambda$, the area of warm color pixels begins to grow, and $\Gini(A_{r}^{f}(x))$ decreases.
	
	To conclude, while the PGDAT-trained model focuses on a few ``robust pixels'', models trained with IGD tend to focus on ``robust regions'' which contain many pixels with high attribution values. The equality within the robust region releases global inequality but may not be effective in releasing regional inequality as important pixels still cluster in the object area.

	\subsection{Robustness against INA}
	\label{sec:INARobustnessResult}
	
	In this section, we analyze the INA-robustness of models trained with different methods. Results are presented in Figure~\ref{fig:allNoise}\subref{imagenet100NoiseA}, Figure~\ref{fig:allNoise}\subref{imagenet100NoiseB}, Figure~\ref{fig:allNoise}\subref{cifar100NoiseA}, and Figure~\ref{fig:allNoise}\subref{cifar100NoiseB}. On ImageNet-100, all dashed lines are far below the orange line, demonstrating that IGD can effectively improve the $\ell_{\infty}$-adversarially trained model's INA-robustness.  When $\lambda>1$, models trained with IGD even have better INA-robustness than the standard-trained model. On CIFAR100, both PGDAT- and IGD-trained models have better INA-robustness than the standard-trained model's, and the improvement of INA-robusstness gained by IGD is minor, which is dissimilar to results on ImageNet-100 and is discussed in detail in Section~\ref{sec:assmption}. On CIFAR100 and ImageNet-100, all IGD- and PGDAT-trained models have a higher $\Gini(A^{f}(x))$ than the standard-trained model, but some of them still have better INA-robustness. These results indicate that a higher $\Gini(A^{f}(x))$ does not necessarily lead to worse INA-robustness, which we discuss in Section~\ref{sec:giniRelation}.

	\subsection{Robustness against IOA}
	\label{sec:occRobustnessResult}
	In this section, we discuss the IOA-robustness of models trained with different methods. Results are presented in Table~\ref{tab:ioaRateTable}. On CIFAR100, both PGDAT- and IGD-trained models have better robustness against IOA than the standard-trained model, and the difference in IOA-robustness between the PGDAT-trained model and the IGD-trained model is not notable. On ImageNet-100, the PGDAT-trained model has worse IOA-robustness than the standard-trained model, which is consistent with~\citet{duaninequality} results. With $2 \le \lambda \le 4$, IGD-trained models have notably better robustness against IOA than the PGDAT-trained model's. We also notice that IOA-robustness gained by IGD with $\lambda=1$ is minor, which we discuss in Section~\ref{sec:giniRelation}. Another interesting finding is that IOA-robustness gained by IGD has better generalization than that gained by CutOut. Compared to the PGDAT-trained model, the PGDAT-trained model using CutOut has competitive robustness against IOA-B but has similar or even worse robustness against IOA-G and IOA-W, which suggests that there exist other factors that influence IOA-robustness since both IOA-B and CutOut occlude images with black patches.

		\begin{table}[!t]
		\centering
		\scriptsize
		\centering
		\caption{Error rates $\downarrow$ of ResNet18 across different methods and types of IOA on CIFAR100 and ImageNet-100. The lowest error rate in each column is \textbf{bold}.}
		\label{tab:ioaRateTable}
		\resizebox{\linewidth}{!}{
			\begin{tabular}{@{}ccccc@{}}
				\toprule
				Dataset &Method           & IOA-B   & IOA-G   & IOA-W   \\ \midrule
				\multirow{7}{*}{CIFAR100} &Standard         & 27.37\% & 17.84\% & \textbf{32.86\%} \\
				&PGDAT\cite{PGD}            & 24.00\% & \textbf{8.93\%}  & 34.67\% \\
				&PGDAT+CutOut\cite{duaninequality}     & \textbf{7.37\%}  & 10.29\% & 35.78\% \\
				&IGD ($\lambda$=1) & 23.02\% & 9.93\%  & 33.66\% \\
				&IGD ($\lambda$=2) & 23.94\% & 9.29\%  & 33.80\% \\
				&IGD ($\lambda$=3) & 25.55\% & 10.23\% & 35.18\% \\
				&IGD ($\lambda$=4) & 25.47\% & 10.31\% & 36.12\% \\ \midrule
				\multirow{7}{*}{ImageNet-100} &Standard         & \textbf{17.59\%} & 8.25\%  & \textbf{13.71\%} \\
				&PGDAT\cite{PGD}            & 32.31\% & 10.85\% & 40.76\% \\
				&PGDAT+CutOut\cite{duaninequality}     & 21.17\% & 10.12\% & 39.94\% \\
				&IGD ($\lambda$=1) & 31.49\% & 10.22\% & 38.30\% \\
				&IGD ($\lambda$=2) & 22.68\% & 5.98\%  & 28.53\% \\
				&IGD ($\lambda$=3) & 19.82\% & \textbf{5.19\%}  & 24.62\% \\
				&IGD ($\lambda$=4) & 18.01\% & 5.46\%  & 24.23\% \\ \bottomrule
			\end{tabular}
		}
	\end{table}
	
	\begin{table}[!t]
		\centering
		\caption{Error rates $\downarrow$ of different methods against ImageNet-C's Gaussian noise. Level $n$ represents the severity of the noise, and a larger $n$ means a more severe noise. The lowest error rate in each column is \textbf{bold}.}
		\label{tab:imagenetC_Gaussian}
		\resizebox{\linewidth}{!}{
			\begin{tabular}{@{}cccccc@{}}
				\toprule
				Method   & Level 1       & Level 2       & Level 3       & Level 4       & Level 5       \\ \midrule
				Standard         & 15.19\% & 35.50\% & 66.31\% & 88.23\% & 96.45\% \\
				PGDAT\cite{PGD}            & 5.06\%  & 14.07\% & 38.17\% & 69.43\% & 91.98\% \\
				PGDAT+CutOut\cite{duaninequality}    & 4.54\%  & 14.73\% & 42.37\% & 74.33\% & 93.26\% \\
				IGD ($\lambda$=1) & 3.02\%  & 10.06\% & 28.47\% & 60.32\% & 88.79\% \\
				IGD ($\lambda$=2) & 3.42\%  & 9.30\%  & 26.07\% & 56.67\% & 86.92\% \\
				IGD ($\lambda$=3) & \textbf{2.86\%}  & \textbf{8.25\%}  & 24.03\% & 52.93\% & 85.44\% \\
				IGD ($\lambda$=4) & 3.62\%  & 9.50\%  & \textbf{23.64\%} & \textbf{51.31\%} & \textbf{82.74\%} \\ \bottomrule
		\end{tabular}}
	\end{table}
	\begin{table}[!t]
		\centering
		\caption{Error rates $\downarrow$ of different methods against ImageNet-C's impulse noise. Level $n$ represents the severity of the noise, and a larger $n$ means a more severe noise. The lowest error rate in each column is \textbf{bold}.}
		\label{tab:imagenetC_Impulse}
		\resizebox{\linewidth}{!}{
			\begin{tabular}{@{}cccccc@{}}
				\toprule
				Method            & Level 1 & Level 2 & Level 3 & Level 4 & Level 5 \\ \midrule
				Standard         & 34.45\% & 57.53\% & 72.65\% & 90.60\% & 96.09\% \\
				PGDAT\cite{PGD}            & 13.77\% & 33.43\% & 53.19\% & 81.20\% & 94.61\% \\
				PGDAT+CutOut\cite{duaninequality}     & 14.73\% & 37.84\% & 56.71\% & 84.62\% & 95.00\%    \\
				IGD ($\lambda$=1) & 9.99\%  & 24.95\% & 40.86\% & 73.60\% & 91.35\% \\
				IGD ($\lambda$=2) & 7.69\%  & 21.83\% & 37.57\% & 69.79\% & 90.43\% \\
				IGD ($\lambda$=3) & \textbf{6.21\%}  & \textbf{18.34\%} & \textbf{32.08\%} & 65.38\% & 88.82\% \\
				IGD ($\lambda$=4) & 7.50\%  & 19.76\% & 32.71\% & \textbf{63.25\%} & \textbf{85.93\%} \\ \bottomrule
		\end{tabular}}
	\end{table}
	\begin{table}[!t]
		\centering
		\caption{Error rates $\downarrow$ of different methods against ImageNet-C's shot noise. Level $n$ represents the severity of the noise, and a larger $n$ means a more severe noise. The lowest error rate in each column is \textbf{bold}.}
		\label{tab:imagenetC_shot}
		\resizebox{\linewidth}{!}{
			\begin{tabular}{@{}cccccc@{}}
				\toprule
				Method            & Level 1 & Level 2 & Level 3 & Level 4 & Level 5 \\ \midrule
				Standard         & 18.28\% & 39.78\% & 65.78\% & 88.53\% & 94.25\% \\
				PGDAT\cite{PGD}             & 7.65\%  & 20.68\% & 45.13\% & 77.22\% & 90.11\% \\
				PGDAT+CutOut\cite{duaninequality}     & 6.96\%  & 21.79\% & 49.05\% & 81.33\% & 91.72\% \\
				IGD ($\lambda$=1) & 4.80\%  & 14.83\% & 33.89\% & 70.02\% & 86.16\% \\
				IGD ($\lambda$=2) & 4.67\%  & 14.07\% & 31.20\% & 66.47\% & 83.83\% \\
				IGD ($\lambda$=3) & \textbf{3.91\%}  & \textbf{12.36\%} & 29.36\% & 65.15\% & 83.73\% \\
				IGD ($\lambda$=4) & 4.34\%  & 13.31\% & \textbf{29.16\%} & \textbf{62.00\%}    & \textbf{81.07\%} \\ \bottomrule
		\end{tabular}}
	\end{table}

	\subsection{Robustness against i.i.d. random noise}
	\label{sec:imagenetCRobustness}
	Apart from inductive attacks introduced in Section~\ref{sec:introInequal}, we use RN~\cite{duaninequality} to evaluate the robustness of models trained on CIFAR100 and ImageNet-100. We also use noisy images in the subset of ImageNet-C~\cite{imagenetC}, which has the same classes as ImageNet-100's, to benchmark the robustness of models trained on ImageNet-100 against common noises. The main difference between noisy images of ImageNet-C and RN is that the variance of the random noise in ImageNet-C is smaller than that of RN. Results on RN are presented in Figure~\ref{fig:allNoise}\subref{imagenet100NoiseC} and Figure~\ref{fig:allNoise}\subref{cifar100NoiseC}. Results on noisy images of ImageNet-C are shown in Table~\ref{tab:imagenetC_Gaussian}, Table~\ref{tab:imagenetC_Impulse}, and Table~\ref{tab:imagenetC_shot}.
	
	In Figure~\ref{fig:allNoise}\subref{imagenet100NoiseC}, we find that, on ImageNet-100, the PGDAT-trained model is still more vulnerable to RN than the standard-trained model even though pixels perturbed by noise are randomly selected. This result can be intriguing since~\citet{otherfactor1} claimed that adversarial training biases the model towards low-frequency information in the input and makes the model robust to high-frequency perturbation, like noises. In comparison, IGD can promote $\ell_{\infty}$-adversarially trained model's robustness against RN, with error rates dropping from around 65\% to 10\%. On CIFAR100, the RN-robustness gained by IGD is less notable than that on ImageNet-100, and the PGDAT-trained model has better RN-robustness than the standard-trained model, which is opposite to results on ImageNet-100. Such differences in robustness among datasets with different resolutions also exist when models are attacked by INA or IOA, which we explain in Section~\ref{sec:assmption}.
	
	As for results on ImageNet-C, all IGD-trained models have lower error rates than the standard-trained model's and the PGDAT-trained model's. Compared to PGDAT, IGD can reduce the error rate of the models by up to 21.11\%. We also find that CutOut worsens the PGDAT-trained model's robustness against noises and increases the PGDAT-trained model's error rate by up to 4.9\%. Unlike the results of RN, the standard-trained model has worse robustness against noisy images in ImageNet-C than the PGDAT-trained model, which we discuss in Section~\ref{sec:explain}.

	\section{Discussion and Analysis}
	\label{sec:disscuss}
	In this section, we discuss and analyze the factors that influence the model's inequality-based robustness.  In Section~\ref{sec:explain}, we reduce the analysis of inequality-based robustness to the analysis of input gradients and explain the relationship between the Gini value and the deviation of the class score, which bridges the gap between the inequality phenomenon and the model's robustness against noise and occlusion. In Section~\ref{sec:assmption}, we conjecture the reasons why the inequality phenomenon and its threats are not apparent on low-resolution datasets like CIFAR100~\cite{cifar}. In Section~\ref{sec:ablation}, we conduct ablation studies to explore different attribution methods, the compatibility of IGD with other adversarial training methods, the adversarial-inequality trade-off, and the effects of each component in IGD. All experiments in Section~\ref{sec:ablation} and Section~\ref{sec:explain} are conducted on the high-resolution dataset ImageNet-100 for a better illustration.

	\subsection{Theoretical analysis of inequality-based robustness}
	\label{sec:explain}
	In this section, we conduct a theoretical analysis to explain why models with equal input gradients tend to have better inequality-based robustness. The main idea is to reduce the analysis of the inequality-based robustness to the analysis of the input gradients and analyze how the distribution of the input gradients influences the model's inequality-based robustness.
	\subsubsection{\textbf{Preliminary}} \label{sec:preliminary}
	
	For simplicity, we consider a linear score model $f(\vec{x}) = \vec{w}^{\mathrm{T}}\vec{x} + b$, where $\vec{w} \in \mathbb{R}^{n \times 1}$ is the input gradient, $\vec{x} \in \mathbb{R}^{n \times 1}$ is the input, and $b \in \mathbb{R}$ is a bias. For a more complicated non-linear model like the CNN, following \citet{saliency}, we can approximate the non-linear model as a linear score model by computing first-order Taylor expansion: $f^{y}(\vec{x}) \approx (\frac{\partial f^{y}(\vec{x})}{\partial \vec{x}})^{\mathrm{T}}\vec{x} + b$. The input $\vec{x}$ is perturbed by noise $\vec{\delta} \in \mathbb{R}^{n \times 1}$ ($\delta_{i} \sim \mathcal{D}(\mu_{\delta}, \sigma_{\delta}^{2})$) masked by $\vec{m} \in \left\{0, 1\right\}^{n \times 1}$, where $m_{i}$ stands for the $i$th component of $\vec{m}$, and $\mathcal{D}(\mu, \sigma^{2})$ represents some distribution with a mean $\mu$ and a variance $\sigma^{2}$. Assuming $k$ pixels are perturbed, formally, we have
	\begin{align}
		m_{i} &= \begin{cases}
			1,\; &i \in \left\{d_1, d_2, ... , d_k\right\};\\
			0,\; &\text{otherwise}.
		\end{cases}
	\end{align}
	We denote $\vec{w}^k$ as $\left[w_{d_1}, w_{d_2}, \ldots , w_{d_{k}}\right]^{\mathrm{T}}$ in the rest of Section~\ref{sec:disscuss}.

	Our theoretical analysis is conducted under two constraints. One is that $\left\|\vec{w}\right\|_{1}$ is fixed because we intend to analyze how the Gini value is correlated with the inequality-based robustness, and because the Gini value is determined by components' relative value rather than their absolute value. Another is that the clean example's class score $y$ is fixed. The reason is that the additivity of the linear score model we consider allows us to disentangle the perturbed term from the clean class score (see \cref{eq:linear}). Since IGD can well control $\left\|\vec{w}\right\|_{1}$ and $y$ (see Table~\ref{tab:advGinitable} and Table~\ref{tab:sumOfAbswAndConf}), we can approximately assume these two constraints hold and apply the results of our theoretical analysis (\cref{sec:theoretical}) under such constraints to the empirical analysis (\cref{sec:empirical}) we conduct.
	
	\subsubsection{\textbf{The correlation between the input gradients and the robustness against noise}}\label{sec:theoretical}
	We test two types of noise in our paper. One is the additive noise like INA1, Gaussian noise, and shot noise, where $\vec{x}' = \vec{x} + \vec{\delta}*\vec{m}$, and the other is the multiplicative-additive noise like INA2 and impulse noise, where $\vec{x}' = \vec{x} * (1 - \vec{m}) + \vec{\delta} * \vec{m}$. For conciseness, we also view occlusion as a noise whose formula of $\vec{x}'$ is the same as the multiplicative-additive noise's. In this section, we explain how we analyze the model's robustness against noise through the input gradients.\\
	
	\noindent\textbf{Case of additive noise.}\  When the input perturbed by additive noise is fed into the model, we have
	\begin{align}\nonumber
		y' &= \vec{w}^{\mathrm{T}}(\vec{x}+\vec{\delta}*\vec{m}) + b\\
		&= y + \vec{w}^{\mathrm{T}}(\vec{\delta}*\vec{m}) = y + y_{\delta}\,, \label{eq:linear}
	\end{align}
	where $y_{\delta} \sim \mathcal{D}(\mu_{\delta}\sum_{i=1}^{k}w_{d_{i}}, \sigma_{\delta}^{2}\sum_{i=1}^{k}w_{d_{i}}^{2})$, and $y$ is the clean example's class score. To make the model robust against $\vec{\delta}$, we need to suppress $y_{\delta}$'s impact by enlarging $y$ to submerge $y_{\delta}$ or suppressing $y_{\delta}$. We mainly discuss the latter one, where we suppress the deviation of class score
	\begin{equation}
		\label{eq:err}
		\mathbb{E}\left\{(y' - y)^{2}\right\} = \mathbb{E}\left\{y_{\delta}^{2}\right\} =  \mu_{\delta}^{2}(\sum_{i=1}^{k}w_{d_{i}})^2 + \sigma_{\delta}^{2}\sum_{i=1}^{k}w_{d_{i}}^{2}\,.
	\end{equation}
	Consequently, we can analyze the model's noise-robustness by comparing $(\sum_{i=1}^{k}w_{d_{i}})^2$ and $\sum_{i=1}^{k}w_{d_{i}}^{2}$ which are both determined by the input gradients.\\
	
	\noindent\textbf{Case of Multiplicative-additive noise and occlusion.}\  We can transform the multiplicative-additive noise and the occlusion into the form of additive noise: $\vec{x}' = \vec{x} + (\delta - \vec{x}) * \vec{m}$. Formally, given $x \sim \mathcal{D}(\mu_{x}, \sigma_{x}^{2})$, we have the deviation of class score
	\begin{align}
		\mathbb{E}\left\{(y' - y)^{2}\right\} = (\mu_{\delta} - \mu_{x})^{2}(\sum_{i=1}^{k}w_{d_{i}})^2 + (\sigma_{\delta}^{2}+\sigma_{x}^{2})\sum_{i=1}^{k}w_{d_{i}}^{2}
	\end{align}
	which is identical to Eqn~(\ref{eq:err}) with different coefficient of $(\sum_{i=1}^{k}w_{d_{i}})^2$ and $\sum_{i=1}^{k}w_{d_{i}}^{2}$.
	
	\subsubsection{\textbf{The relationship between the Gini value and the deviation of the class score}} \label{sec:giniRelation} After reducing the analysis of the inequality-based robustness to the analysis of the input gradient, we analyze the relationship between the Gini value of the input gradients and the deviation of the class score which is determined by the input gradients.\\
	
	\noindent\textbf{Decreasing the Gini value without altering the sum of $\Phi$.}\  We first introduce a basic and the only operation that decreases the Gini value without altering the sum of $\Phi$. We fix $\sum_{i=1}^{n}\phi_{i}$ because we make a constraint where $\left\|\vec{w}\right\|_{1}$ is fixed during our analysis.
	
	Recall our notation in Section~\ref{sec:introInequal}. For two components in $\Phi$, say, $\phi_{a}$ and $\phi_{b}$, where $a < b$, the only approach to change their relative values without altering $\sum_{i=1}^{n}\phi_{i}$ is to increase $\phi_{a}$ by $\Delta$ and decrease $\phi_{b}$ by $\Delta$, where $\phi_{b} - \phi_{a} \ge 2*\Delta$. After this operation, we obtain $\Phi^{'} = \left\{\phi^{'}_{i}, i=1...n \mid 0 \le \phi^{'}_{i} \le \phi^{'}_{i+1} \right\}$, having
	\begin{equation}
		\phi^{'}_{i} = \begin{cases}
			\phi_{a} + \Delta, &i = a';\\
			\phi_{b} - \Delta, &i = b';\\
			\phi_{i+1}, &i \in [a, a');\\
			\phi_{i-1}, &i \in (b', b];\\
			\phi_{i}, &\text{otherwise}.
		\end{cases}\,, 
	\end{equation}
	where $a \le a' < b' \le b$, and
	\begin{align}
		\nonumber&\Gini(\Phi^{'}) - \Gini(\Phi)\\ \nonumber&= -\frac{\sum_{i=a+1}^{a'}(\phi_{i} - \phi_{a}) + \sum_{i=b'}^{b-1}(\phi_{b} - \phi_{i}) + (b' - a')\Delta}{(n / 2)*\sum_{i=1}^{n}\phi_{i}}\\
		& < 0\,. \label{eq:deltaGini}
	\end{align}
	Consequently, this operation can decrease the Gini value monotonically without altering $\sum_{i=1}^{n}\phi_{i}$.  We can transfer a vector like $\vec{w}$ into $\Phi$ by taking the absolute value of $\vec{w}$ and sorting it in increasing order, which bridges the gap between the analysis of the Gini value and our analysis of the deviation of the class score.\\

	\noindent\textbf{Case of RN.}\  In this case, since $\vec{w}^k$ can be seen as a random sampling of $\vec{w}$, where $k$ is large enough such that the distribution of $\vec{w}^k$ is similar to the distribution of $\vec{w}$, we have $\Gini(\vec{w}^k) \approx \Gini(\vec{w})$ and that $\left\|\vec{w}^k\right\|_{1}$ can be seen fixed.
	
	For two components $\vec{w}_{a}$ and $\vec{w}_{b}$ in $\vec{w}^k$, if we increase $\left|\vec{w}_{a}\right|$ and decrease $\left|\vec{w}_{b}\right|$ by $\Delta$, where $\left|\vec{w}_{a}\right| < \left|\vec{w}_{b}\right|$ and $\left|\vec{w}_{b}\right| - \left|\vec{w}_{a}\right| \ge 2 * \Delta$, without altering $\left\|\vec{w}^k\right\|_{1}$, we decrease $\Gini(\vec{w}^k)$ and obtain $\vec{w'}^k$ where
	\begin{align}
		\nonumber
		\sum_{i=1}^{k}{w_{d_{i}}^{'}}^{2} &= \sum_{i=1}^{k}w_{d_{i}}^{2} + 2 * \Delta *(\Delta - \left|w_{b}\right| + \left|w_{a}\right|)\\
		&\le \sum_{i=1}^{k}w_{d_{i}}^{2} - 2 * \Delta^2 < \sum_{i=1}^{k}w_{d_{i}}^{2}\,.\label{eq:deltaVar}
	\end{align}
	Consequently, \emph{decreasing $\Gini(\vec{w}^k)$ is actually decreasing $y_{\sigma}$'s variance}. Such a property explains why models with lower $\Gini(\vec{w})$ tend to have better robustness against noise and occlusion.
	
	$(\sum_{i=1}^{k}w_{d_{i}})^2$, however, does not have the same monotonicity as the Gini value like $\sum_{i=1}^{k}w_{d_{i}}^{2}$. After increasing $\left|w_{a}\right|$ and decreasing $\left|w_{b}\right|$ by $\Delta$, $(\sum_{i=1}^{k}w_{d_{i}})^2$ will  vary if $w_{a}$ and $w_{b}$ have different sign. Consequently, we can only analyze $(\sum_{i=1}^{k}w_{d_{i}})^2$ through empirical experiments.\\
	
	\noindent\textbf{Case of INA.}\  Unlike the case of RN where $\vec{w}^k$ can be seen as a random sampling of $\vec{w}$, $\vec{w}^k$ in the case of INA does not have $\Gini(\vec{w}^k) \approx \Gini(\vec{w})$ and that $\left\|\vec{w}^k\right\|_{1}$ can be seen as a constant. Nonetheless, since INA selects pixels with the $k$ largest $w_{i}$ to perturb, the model having a lower Gini value tends to have smaller $\sum_{i=1}^{k}w_{d_{i}}^{2}$ and $(\sum_{i=1}^{k}w_{d_{i}})^2$ when attacked by INA, which can be measured through empirical experiments.

	\subsubsection{\textbf{Empirical experiments for validating theoretical analysis}}\label{sec:empirical} 
\begin{figure}[!t]
		\centering
		\begin{minipage}{\linewidth}
			\centering
			\includegraphics[width=\linewidth]{./experimentsImage/legend7.pdf}\\
			\centering
			\subfloat[$\sum_{i=1}^{k}w_{d_{i}}^{2}$ of RN]{\label{fig:devOnImagenet100}
				\includegraphics[width=0.48\linewidth]{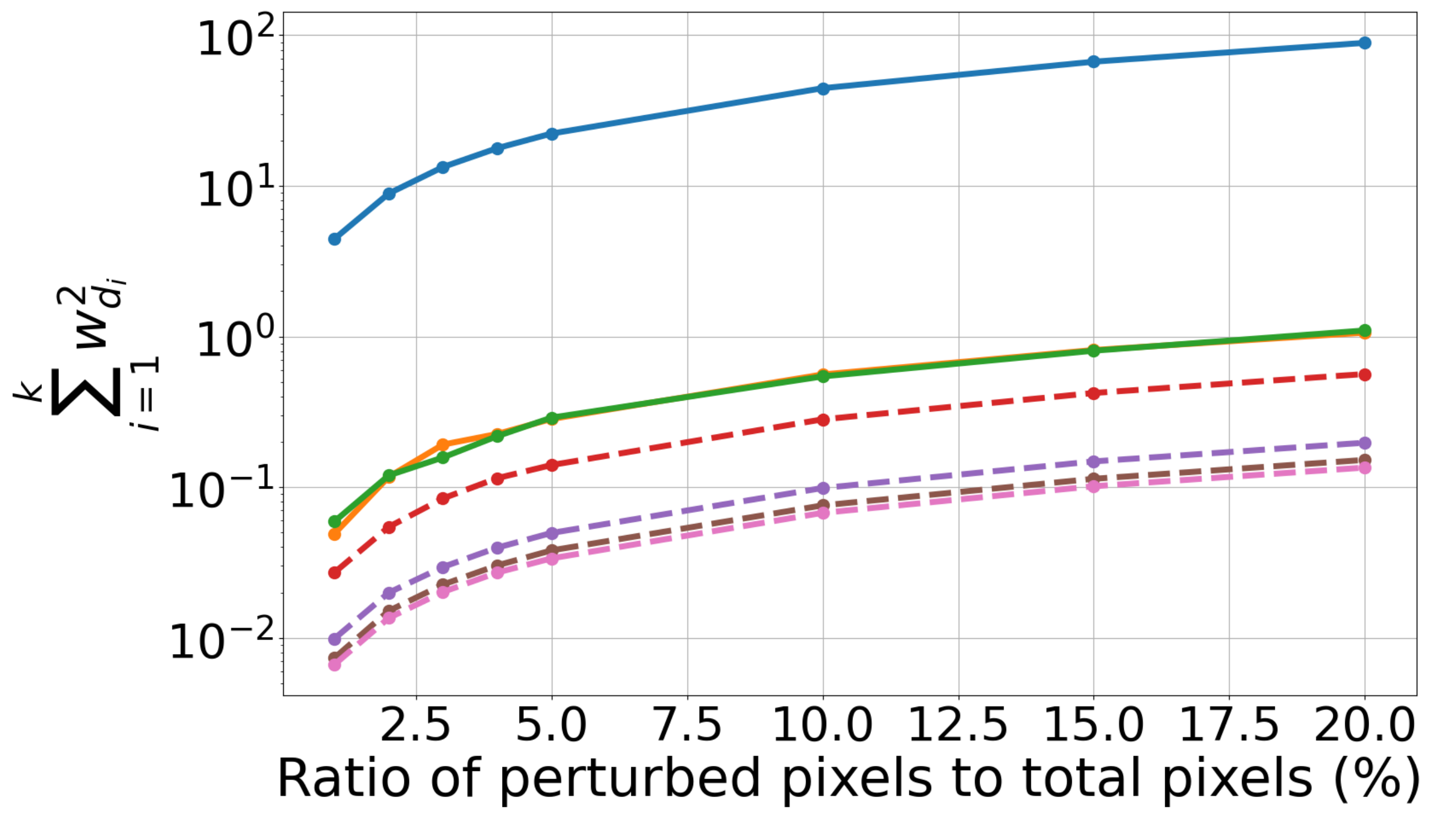}
			}
			\subfloat[$(\sum_{i=1}^{k}w_{d_{i}})^{2}$ of RN]{\label{fig:meanOnImagenet100}
				\includegraphics[width=0.48\linewidth]{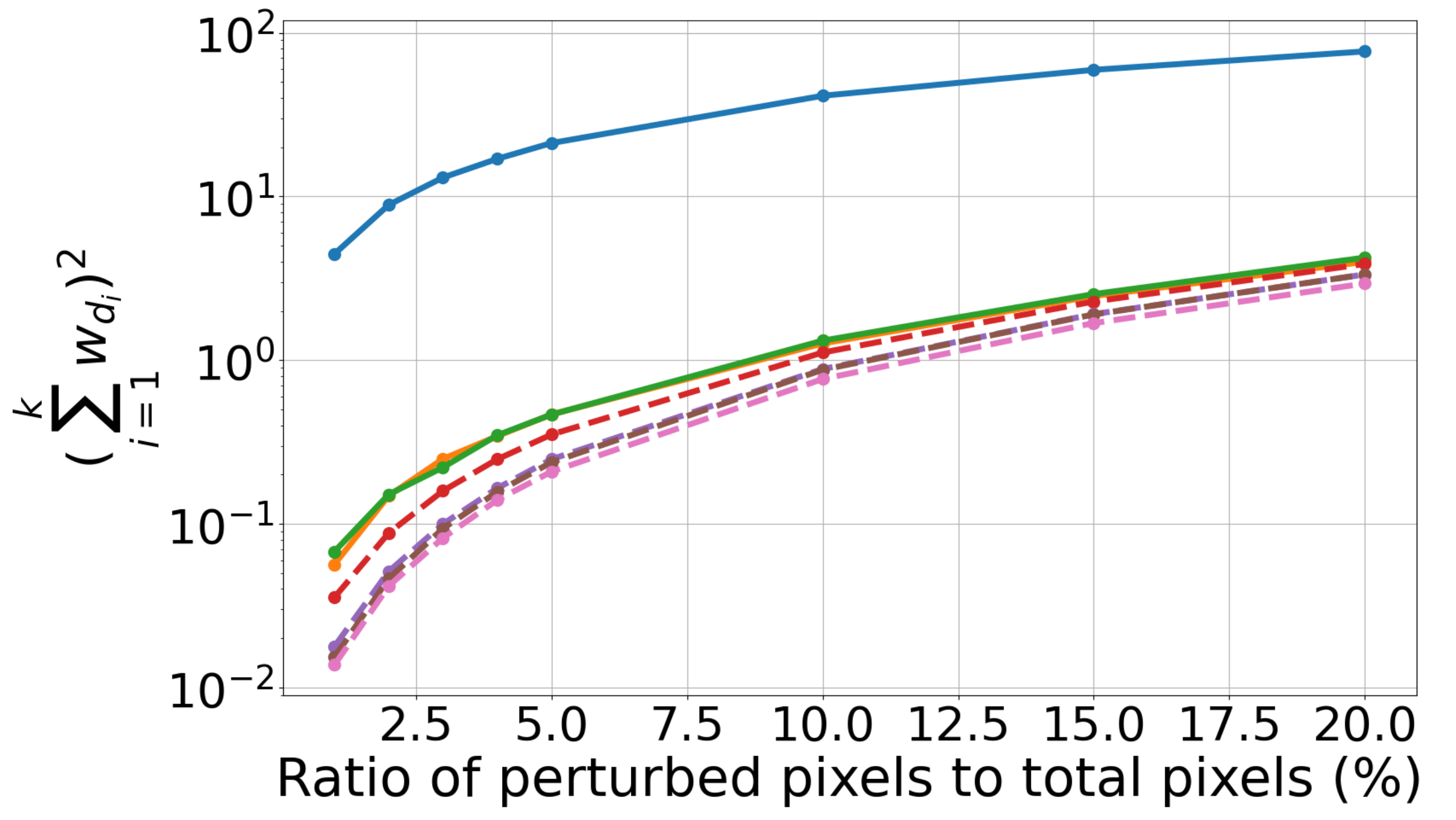}
			}\\
			\subfloat[$\sum_{i=1}^{k}w_{d_{i}}^{2}$ of INA]{\label{fig:INAImagenet100DEV}
				\includegraphics[width=0.48\linewidth]{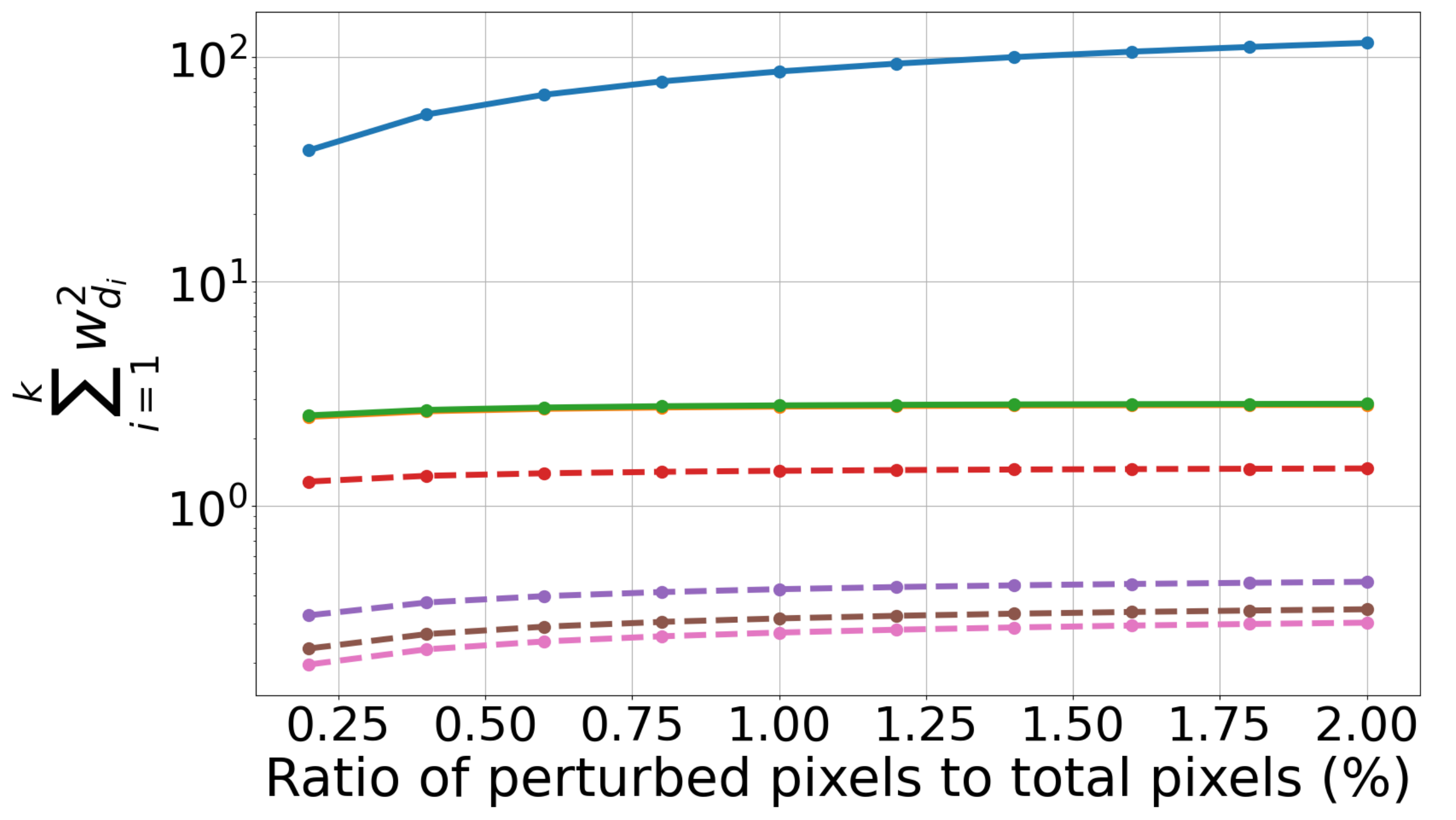}
			}
			\subfloat[$(\sum_{i=1}^{k}w_{d_{i}})^{2}$ of INA]{\label{fig:INAImagenet100MEAN}
				\includegraphics[width=0.48\linewidth]{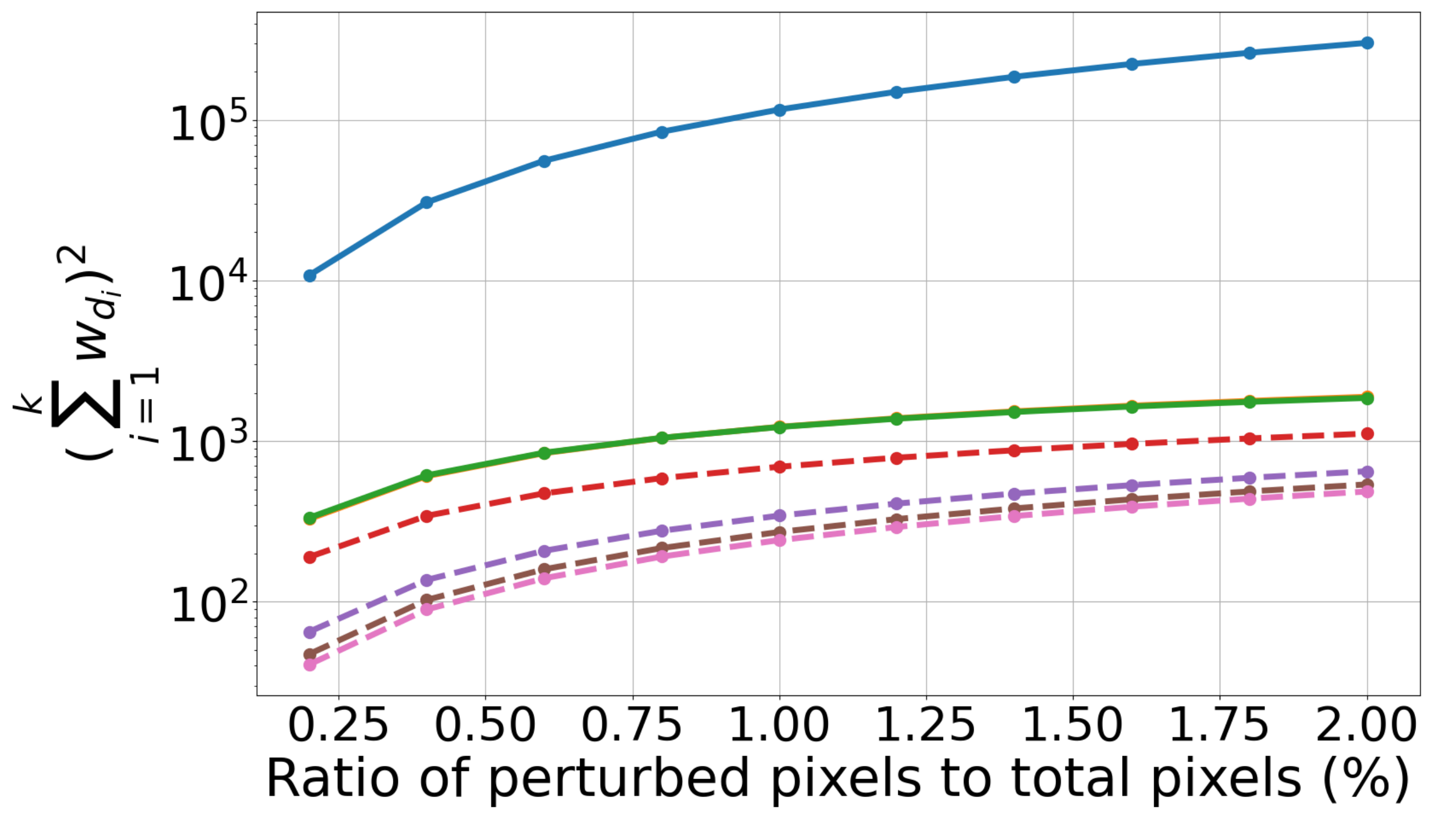}
			}
			
		\end{minipage}
		\caption{Average $\sum_{i=1}^{k}w_{d_{i}}^{2}$ and $(\sum_{i=1}^{k}w_{d_{i}})^{2}$ across different methods on ImageNet-100. We can find that all dashed lines are below solid lines, which indicates a lower deviation of the class score. Best viewed in color.}\label{fig:inputgradient}
	\end{figure}

To further validate our theoretical analysis, on ImageNet-100, we measure $\sum_{i=1}^{k}w_{d_{i}}^{2}$ and $(\sum_{i=1}^{k}w_{d_{i}})^{2}$ of models attacked by RN and INA with different numbers $k$ of perturbed pixels, which are presented in Figure~\ref{fig:inputgradient}\subref{fig:devOnImagenet100}, Figure~\ref{fig:inputgradient}\subref{fig:meanOnImagenet100}, Figure~\ref{fig:inputgradient}\subref{fig:INAImagenet100DEV}, and Figure~\ref{fig:inputgradient}\subref{fig:INAImagenet100MEAN}. We also measure $\sum_{i=1}^{k}w_{d_{i}}^{2}$ and $(\sum_{i=1}^{k}w_{d_{i}})^{2}$ of models attacked by IOA, which is presented in Table~\ref{tab:sumOfAbswAndConf}.\\

	\noindent\textbf{Remark on Figure~\ref{fig:inputgradient}\subref{fig:devOnImagenet100}.}\  The $\sum_{i=1}^{k}w_{d_{i}}^{2}$ of IGD are all smaller than that of PGDAT. This outcome is consistent with our analysis in Section~\ref{sec:giniRelation} because the IGD-trained model has a lower Gini value than the PGDAT-trained model. Moreover, Figure~\ref{fig:inputgradient}\subref{fig:devOnImagenet100} can already explain why the IGD-trained model has better robustness against RN, Gaussian noise, and shot noise. This is because these three types of noise have $\mu_{\delta} = 0$, which suggests that the variance of the class score alone determines the deviation of the class score.\\
	
	\noindent\textbf{Remark on Figure~\ref{fig:inputgradient}\subref{fig:meanOnImagenet100}.}\  Though a lower Gini value does not necessarily mean lower $(\sum_{i=1}^{k}w_{d_{i}})^{2}$, the IGD-trained models still have lower $(\sum_{i=1}^{k}w_{d_{i}})^{2}$. We think this is because $(\sum_{i=1}^{k}w_{d_{i}})^{2} = \sum_{i=1}^{k}w_{d_{i}}^{2} + 2 * \sum_{i=1}^{k}\sum_{j=1}^{k}w_{d_{i}}w_{d_{j}}$ where the gap of $\sum_{i=1}^{k}w_{d_{i}}^{2}$ is large enough to determine the relative value of $(\sum_{i=1}^{k}w_{d_{i}})^{2}$. Results in Figure~\ref{fig:inputgradient}\subref{fig:meanOnImagenet100} can explain why the IGD-trained model has better robustness against impulse noise.\\
	
	\noindent\textbf{Remark on Figure~\ref{fig:inputgradient}\subref{fig:INAImagenet100DEV}-\subref{fig:INAImagenet100MEAN}.}\  We can infer from Figure~\ref{fig:inputgradient}\subref{fig:INAImagenet100DEV} and Figure~\ref{fig:inputgradient}\subref{fig:INAImagenet100MEAN} that, when attacked by INA, all IGD-trained models have lower $(\sum_{i=1}^{k}w_{d_{i}})^{2}$ and $\sum_{i=1}^{k}w_{d_{i}}^{2}$. This result is consistent with our assumption in Section~\ref{sec:giniRelation} since IGD does not alter $\left\|\vec{w}\right\|_{1}$ significantly compared to PGDAT and has a $\vec{w}$ with a more even distribution than PGDAT. Consequently, IGD improves the INA-robustness of the model.\\
	
	\noindent\textbf{Remark on IOA.}\   As mentioned in Section~\ref{sec:releaseInequality}, pixels that are considered important by the IGD-trained model tend to cluster when $\lambda$ is not high. This property may result in high $(\sum_{i=1}^{k}w_{d_{i}})^{2}$  (see IGD ($\lambda$=1) in Table~\ref{tab:sumOfAbswAndConf}) and worse or non-improved IOA-robustness, which is shown in Table~\ref{tab:ioaRateTable}. When $\lambda$ is increased, the relatively low $(\sum_{i=1}^{k}w_{d_{i}})^{2}$ and $\sum_{i=1}^{k}w_{d_{i}}^{2}$ weaken the effect of occlusion, and the IGD-trained models' robustness against all three types of occlusion are improved. Moreover, Unlike the other two types of noise, the variance $\sigma_{\delta}^2$ of IOA is $0$, indicating $\mathbb{E}\left\{(y' - y)^{2}\right\} = (\mu_{\delta} - \mu_{x})^{2}(\sum_{i=1}^{k}w_{d_{i}})^{2} + \sigma_{x}^{2}\sum_{i=1}^{k}w_{d_{i}}^{2}$.  The smaller coefficient of $\sum_{i=1}^{k}w_{d_{i}}^{2}$ further weakens the robustness gained by the equal decision pattern, which explains why IOA-robustness gained by IGD is not as notable as the INA-robustness and the RN-robustness.\\

	\begin{table}[!t]
		\centering
		\caption{Average $\sum_{i=1}^{k}w_{d_{i}}^{2}$ and $(\sum_{i=1}^{k}w_{d_{i}})^{2}$ of ResNet18 attacked by IOA, average confidence of correctly classified example, and logit margin loss~\cite{LML} across different methods on ImageNet-100.}
		\label{tab:sumOfAbswAndConf}
		\resizebox{\linewidth}{!}{
			\begin{tabular}{@{}ccccc@{}}
				\toprule
				Method           & $\sum_{i=1}^{k}w_{d_{i}}^{2}$ & $(\sum_{i=1}^{k}w_{d_{i}})^{2}$ & Confidence & Logit margin loss \\ \midrule
				Standard         & 166.21    & 169.09    & 96.73\%    & -8.56             \\
				PGDAT~\cite{PGD}            & 1.12      & 8.75      & 77.07\%    & -4.66             \\
				PGDAT+CutOut~\cite{duaninequality}    & 1.06      & 9.53      & 77.43\%    & -4.69             \\
				IGD ($\lambda$=1) & 0.76      & 9.44      & 78.60\%    & -4.89             \\
				IGD ($\lambda$=2) & 0.30       & 7.12      & 79.29\%    & -4.94             \\
				IGD ($\lambda$=3) & 0.24      & 6.04      & 79.29\%    & -4.91             \\
				IGD ($\lambda$=4) & 0.21      & 5.73      & 78.50\%    & -4.81             \\ \bottomrule
			\end{tabular}
		}
	\end{table}
	
	\begin{table*}[!t]
		\centering
		\caption{The global Gini value and the regional Gini value of the standard-trained and $\ell_{\infty}$-adversarially trained model on CIFAR10 and TinyImageNet. $l$ is set to 4 for CIFAR10 and is set to 8 for TinyImageNet.}
		\label{tab:lowResolutionGiniTable}
		\resizebox{\linewidth}{!}{
			\begin{tabular}{@{}cccccccc@{}}
				\toprule
				Datset                        & Metric                                & Model        & Integrated Gradients & Input X Graidients & Shapley Value & SmoothGrad & Saliency Map \\ \midrule
				\multirow{4}{*}{CIFAR10}      & \multirow{2}{*}{$\Gini(A^{f}(x))$}     & Std. trained & 0.615                & 0.616              & 0.613    & 0.593      & 0.532        \\
				&                                       & Adv. trained & 0.719                & 0.745              & 0.734    & 0.889      & 0.737        \\ \cmidrule(l){2-8} 
				& \multirow{2}{*}{$\Gini(A_{r}^{f}(x))$} & Std. trained & 0.346                & 0.426              & 0.424    & 0.518      & 0.337        \\
				&                                       & Adv. trained & 0.477                & 0.596              & 0.595    & 0.778      & 0.592        \\ \midrule
				\multirow{4}{*}{TinyImageNet} & \multirow{2}{*}{$\Gini(A^{f}(x))$}     & Std. trained & 0.609                & 0.575              & 0.586    & 0.452      & 0.501        \\
				&                                       & Adv. trained & 0.650                & 0.653              & 0.673    & 0.674      & 0.578        \\ \cmidrule(l){2-8} 
				& \multirow{2}{*}{$\Gini(A_{r}^{f}(x))$} & Std. trained & 0.235                & 0.348              & 0.365    & 0.345      & 0.273        \\
				&                                       & Adv. trained & 0.333                & 0.461              & 0.499    & 0.552      & 0.396        \\ \bottomrule
		\end{tabular}}
	\end{table*}
	
	\noindent\textbf{Remark on the standard-trained model.}\  We find that the standard-trained model has larger $\sum_{i=1}^{k}w_{d_{i}}^{2}$ and $(\sum_{i=1}^{k}w_{d_{i}})^{2}$, which suggests larger deviation of the class score and worse inequality-based robustness. However, results in Section~\ref{sec:Experiment} and Section~\ref{sec:ablation} demonstrate that the standard-trained model has better inequality-based robustness than the PGDAT-trained model on ImageNet-100. We think this is because the standard-trained model correctly classifies clean examples with high confidence (see Table~\ref{tab:sumOfAbswAndConf}). Even though the deviation of the standard-trained model's class score is large, the clean example's class score $y$ is large enough to submerge $y_{\delta}$, allowing the standard-trained model to classify the noisy example correctly. Moreover, we note that the model with high classification confidence tends to have low logit margin loss (LML)
	\begin{equation}
		\mathcal{L}_{LML} = \max_{i \ne t} f^{i}(x) - f^{t}(x)\,, 
	\end{equation}
	where $t$ is the ground truth of the input $x$'s class.~\citet{confidence} claimed that the model with smaller Logit margin loss has fewer potentially misclassified examples, which suggests better robustness against perturbations.
	
	Besides, we find that the $\sum_{i=1}^{k}w_{d_{i}}^{2}$ of the standard-trained model is only about $100$ times larger than the PGDAT-trained model's. In Table~\ref{tab:advGinitable}, the standard-trained model's $\left\|\vec{w}\right\|_{1}$ is about 50 times larger than the PGDAT's. If the standard-trained model had the same distribution of $\vec{w}$ as that of the PGDAT-trained model, its $\sum_{i=1}^{k}w_{d_{i}}^{2}$ would be about $2,500$ times larger than the PGDAT-trained model's. This finding indicates that the equality of the input gradients still plays an important role in decreasing the deviation of the class score and improving the model's inequality-based robustness. Results in Table~\ref{tab:imagenetC_Gaussian}, Table~\ref{tab:imagenetC_Impulse}, and Table~\ref{tab:imagenetC_shot} can support our claim. As we introduce in Section~\ref{sec:imagenetCRobustness}, the variance of noise in RN is larger than that of ImageNet-C, which suggests that the standard-trained model's robustness against noise is better than that of the PGDAT-trained model if the variance of the noise is large enough. Since the variance of the noise is the coefficient of $\sum_{i=1}^{k}w_{d_{i}}^{2}$ and thus controls the importance of the input gradient, it is the equality of the input gradients that improves the inequality-based robustness of the standard-trained model to be better than the PGDAT-trained model's.

	\subsubsection{\textbf{Conclusion}} The equality of the input gradients gains robustness by suppressing the class score's deviation caused by input space perturbations, and high correct classification confidence further boosts the standard-trained model's robustness.

	\begin{table}[!t]
		\centering
		\caption{Average confidence, $\Gini(A^{f}(x))$, and $\left\|\frac{\partial f^{y}(x)}{\partial x}\right\|_{1}$ of ResNet18 across different methods on CIFAR10.}
		\label{tab:cifar10attrTable}
		\begin{tabular}{@{}cccc@{}}
			\toprule
			Method           & Confidence & $\Gini(A^{f}(x))$     & $\left\|\frac{\partial f^{y}(x)}{\partial x}\right\|_{1}$      \\ \midrule
			Standard           & 99.42\%    & 0.534 & 1,217.57   \\
			PGDAT~\cite{PGD}           & 81.98\%    & 0.702 & 59.10   \\
			PGDAT+CutOut~\cite{duaninequality}           & 77.89\%    & 0.702 & 55.53   \\
			IGD ($\lambda$=1) & 81.37\%    & 0.684 & 42.88   \\
			IGD ($\lambda$=2) & 80.50\%    & 0.644 & 30.81   \\
			IGD ($\lambda$=3) & 76.00\%    & 0.625 & 25.80   \\ \bottomrule
		\end{tabular}
		
	\end{table}
	\begin{table}[!t]
		\centering
		\caption{Error rates $\downarrow$ of ResNet18 trained with different methods on CIFAR10 and TinyImageNet. 10\% of pixels are perturbed. The lowest error rate in each column is \textbf{bold}.}
		\label{tab:cifar10INATable}
		\begin{tabular}{@{}ccccc@{}}
			\toprule
			Dataset                       & Method            & INA1    & INA2    & RN      \\ \midrule
			\multirow{6}{*}{CIFAR10}     & Standard          & 56.05\% & 59.62\% & 76.40\% \\
			& PGDAT             & \textbf{10.11\%} & 24.50\% & \textbf{16.30\%} \\
			& PGDAT+CutOut      & 11.10\% & \textbf{23.24\%} & 19.41\% \\
			& IGD ($\lambda$=1) & 12.42\% & 27.72\% & 16.43\% \\
			& IGD ($\lambda$=2) & 11.38\% & 23.26\% & 15.19\% \\
			& IGD ($\lambda$=3) & 16.61\% & 29.26\% & 18.37\% \\ \midrule
			\multirow{6}{*}{TinyImageNet} & Standard          & 43.39\% & 39.43\% & 38.99\% \\
			& PGDAT             & 23.65\% & 17.86\% & 17.54\% \\
			& PGDAT+CutOut      & 24.53\% & 20.36\% & 19.70\% \\
			& IGD ($\lambda$=1) & 19.95\% & 14.17\% & 14.37\% \\
			& IGD ($\lambda$=2) & 19.95\% & 16.04\% & 15.87\% \\
			& IGD ($\lambda$=3) & \textbf{15.89\%} & \textbf{10.00\%} & \textbf{9.65\%}  \\ \bottomrule
		\end{tabular}
		
	\end{table}

	\subsection{Conjecture on low-resolution dataset}
	\label{sec:assmption}
	On low-resolution datasets like CIFAR100, though the standard-trained model has an equal decision pattern, it does not have better inequality-based robustness, and inequality-based robustness gained by IGD is less notable than that on the high-resolution dataset like ImageNet-100. We attribute this phenomenon to three reasons.
	
	First, the gap of $\Gini(A^{f}(x))$ between the standard- and the $\ell_{\infty}$-adversarially trained model is smaller on low-resolution datasets than on the high-resolution dataset ImageNet-100, which suggests that the gap of $\Gini(A^{f}(x))$ is not large enough to make the $\ell_{\infty}$-adversarially trained model's inequality-based robustness worse than that of the standard-trained model. This conjecture can be supported by results in Section~\ref{sec:Experiment}, where the IGD-trained model has higher $\Gini(A^{f}(x))$ as well as comparable or better inequality-based robustness than the standard-trained model. Intriguingly, we find that, on low-resolution datasets, it is common that the gap of $\Gini(A^{f}(x))$ is small. We evaluate $\Gini(A^{f}(x))$ and $\Gini(A_{r}^{f}(x))$ of ResNet50 trained on CIFAR10 by ~\cite{robustness} and ResNet18 trained on TinyImageNet~\cite{imagenet} by us. Results are reported in Table~\ref{tab:lowResolutionGiniTable}, where we find that the gaps of $\Gini(A^{f}(x))$ on low-resolution datasets are all smaller than that on ImageNet-100.
	
	Second, we hypothesize that the severity of the inequality phenomenon is positively correlated with the number $k$ of pixels. In Section~\ref{sec:giniRelation}, we prove that the Gini value has the same monotonicity as the variance of the class score. After changing $\left|w_{a}\right|$ and $\left|w_{b}\right|$ by $\Delta$, we have
	\begin{equation}
		\frac{\sum_{i=1}^{k}{w_{d_{i}}^{'}}^{2} - \sum_{i=1}^{k}w_{d_{i}}^{2}}{\Gini(\vec{w'}^{k}) - \Gini(\vec{w}^k)} \propto k
	\end{equation}
	This indicates that a larger $k$ leads to a larger descent in the variance of the class score when the gap of the Gini value is fixed and that the importance of the inequality phenomenon grows with the number of pixels. Results in Figure~\ref{fig:allNoise}\subref{imagenet100NoiseC} can support our claim. When $k$ is small, the IGD-trained model with $\lambda=2$ has higher $\Gini(A^{f}(x))$ as well as lower error rates than the standard-trained model. However, as $k$ grows, the standard-trained model tends to have better RN-robustness than the IGD-trained model, which suggests that the inequality phenomenon is severe enough to alter the model's relative robustness. Also, by referring to \cref{tab:cifar10INATable} and \cref{tab:lowResolutionGiniTable}, we can find that the Gini Value gap between the standard-trained model and the adversarially trained model on TinyImageNet, whose resolution ($64\times64$) is smaller than the ImageNet ($256\times256$) but larger than CIFAR ($32\times32$), is at a similar level to that on CIFAR100 and CIFAR10. Yet, the inequality-based robustness gained by the IGD is more stable and significant on TinyImageNet. These results indicate that the importance of the inequality phenomenon grows with the number of pixels.
	
	Third, we acknowledge that the equality of decision pattern and classification confidence are not the only two factors that influence the model's robustness against noise and occlusion. In Table~\ref{tab:sumOfAbswAndConf}, though PGDAT-trained model with CutOut has larger $\sum_{i=1}^{k}w_{d_{i}}^{2}$ and $(\sum_{i=1}^{k}w_{d_{i}})^{2}$, and has higher error rate on IOA-G and IOA-W, its error rates on IOA-B is much lower than IGD ($\lambda$=1, 2). In Table~\ref{tab:cifar10attrTable}, on CIFAR10, the IGD-trained models have lower $\Gini(A^{f}(x))$ and $\left\|\frac{\partial f^{y}(x)}{\partial x}\right\|_{1}$ which means lower $\sum_{i=1}^{k}w_{d_{i}}^{2}$ and $(\sum_{i=1}^{k}w_{d_{i}})^{2}$, but this does not guarantee better inequality-based robustness (see Table~\ref{tab:cifar10INATable}). These two results suggest that there are other factors, like the frequency of perturbation~\cite{otherfactor1}, that may influence the model's inequality-based robustness. These factors may overwhelm the equality of decision pattern and classification confidence when the dataset's resolution is low, or when the gap of the Gini value is small.
	
	To conclude, the unsatisfactory inequality-based robustness of the IGD- and the standard-trained model on CIFAR is due to the intrinsic properties of the low-resolution dataset where the inequality phenomenon and its threats are not severe enough to overwhelm other factors.

	\begin{table*}[!t]
		\centering
		\caption{The global Gini value and the regional Gini value across different datasets, training methods, and attribution methods.}
		\resizebox{\textwidth}{!}{
			\begin{tabular}{@{}cccccccccc@{}}
				\toprule
				\multirow{2}{*}{Dataset}      & \multirow{2}{*}{Method} & \multicolumn{2}{c}{Integrated Gradients} & \multicolumn{2}{c}{Input X Gradient}   & \multicolumn{2}{c}{Shapley Value}            & \multicolumn{2}{c}{SmoothGrad}          \\ \cmidrule(l){3-10} 
				&                         & $\Gini(A^{f}(x))$  & $\Gini(A_{r}^{f}(x))$ & $\Gini(A^{f}(x))$ & $\Gini(A_{r}^{f}(x))$ & $\Gini(A^{f}(x))$ & $\Gini(A_{r}^{f}(x))$ & $\Gini(A^{f}(x))$ & $\Gini(A_{r}^{f}(x))$ \\ \midrule
				\multirow{7}{*}{CIFAR100}     & Standard                & 0.611             & 0.304                & 0.581            & 0.412                & 0.587            & 0.424                & 0.388            & 0.320                \\
				& PGDAT\cite{PGD}                  & 0.672             & 0.377                & 0.702            & 0.569                & 0.696            & 0.584                & 0.640            & 0.554                \\
				& PGDAT+CutOut\cite{duaninequality}           & 0.673             & 0.343                & 0.701            & 0.568                & 0.693            & 0.580                 & 0.605            & 0.508                \\
				& IGD ($\lambda$=1)        & 0.676             & 0.381                & 0.697            & 0.565                & 0.698            & 0.588                & 0.639            & 0.559                \\
				& IGD ($\lambda$=2)        & 0.655             & 0.367                & 0.677            & 0.551                & 0.674            & 0.568                & 0.617            & 0.543                \\
				& IGD ($\lambda$=3)        & 0.642             & 0.362                & 0.660            & 0.534                & 0.660            & 0.557                & 0.607            & 0.542                \\
				& IGD ($\lambda$=4)        & 0.636             & 0.361                & 0.648            & 0.523                & 0.650            & 0.548                & 0.607            & 0.543                \\ \midrule
				\multirow{7}{*}{ImageNet-100} & Standard                & 0.633             & 0.285                & 0.605            & 0.407                & 0.610            & 0.413                & 0.564            & 0.502                \\
				& PGDAT\cite{PGD}                  & 0.936             & 0.382                & 0.942            & 0.601                & 0.947            & 0.633                & 0.941            & 0.594                \\
				& PGDAT+CutOut\cite{duaninequality}           & 0.923             & 0.350                 & 0.935            & 0.596                & 0.941            & 0.624                & 0.933            & 0.562                \\
				& IGD ($\lambda$=1)        & 0.829             & 0.399                & 0.854            & 0.592                & 0.865            & 0.627                & 0.915            & 0.696                \\
				& IGD ($\lambda$=2)        & 0.739             & 0.395                & 0.772            & 0.589                & 0.791            & 0.625                & 0.815            & 0.698                \\
				& IGD ($\lambda$=3)        & 0.715             & 0.375                & 0.744            & 0.573                & 0.764            & 0.609                & 0.807            & 0.692                \\
				& IGD ($\lambda$=4)        & 0.701             & 0.365                & 0.735            & 0.571                & 0.754            & 0.606                & 0.796            & 0.700                \\ \bottomrule
		\end{tabular}}
		\label{tab:otherGiniTable}
	\end{table*}

	\begin{table*}[!t]
		\centering
		\caption{Error rates $\downarrow$ of models across different methods, datasets, and types of INA guided by different attribution methods. 1\% of pixels are perturbed. The lowest error rate in each column is \textbf{bold}.}\label{tab:otherINA}
		\begin{tabular}{@{}cccccccccc@{}}
			\toprule
			\multirow{2}{*}{Dataset}      & \multirow{2}{*}{Method} & \multicolumn{2}{c}{Integrated Gradients} & \multicolumn{2}{c}{Shapley Value} & \multicolumn{2}{c}{Input X Gradients} & \multicolumn{2}{c}{SmoothGrad} \\ \cmidrule(l){3-10} 
			&                         & INA1                & INA2               & INA1            & INA2            & INA1              & INA2              & INA1           & INA2          \\ \midrule
			\multirow{7}{*}{CIFAR100}     & Standard                & 65.43\%             & 77.56\%            & 44.49\%         & 63.53\%         & 44.40\%           & 62.01\%           & 47.58\%        & 52.59\%       \\
			& PGDAT~\cite{PGD}                  & 29.27\%             & 41.36\%            & \textbf{13.13\%}         & \textbf{26.66\%}         & \textbf{10.64\%}           & 21.58\%           & 30.79\%        & 35.16\%       \\
			& PGDAT+CutOut~\cite{duaninequality}            & 33.12\%             & 41.67\%            & 14.36\%         & 28.58\%         & 13.11\%           & 24.15\%           & 33.76\%        & 36.75\%       \\
			& IGD ($\lambda$=1)        & 29.58\%             & 43.19\%            & 13.27\%         & 27.59\%         & 11.47\%           & 21.98\%           & 29.68\%        & 35.92\%       \\
			& IGD ($\lambda$=2)        & 25.47\%             & 38.58\%            & 13.77\%         & 27.61\%         & 11.32\%           & \textbf{20.92\%}           & 26.50\%        & 32.64\%       \\
			& IGD ($\lambda$=3)        & 22.72\%             & 38.03\%            & 15.74\%         & 30.67\%         & 12.18\%           & 24.55\%           & \textbf{25.09\%}        & \textbf{32.34\%}       \\
			& IGD ($\lambda$=4)        & \textbf{20.85\%}             & \textbf{37.25\%}            & 16.05\%         & 32.88\%         & 13.01\%           & 26.92\%           & 25.62\%        & 34.50\%       \\ \midrule
			\multirow{7}{*}{ImageNet-100} & Standard                & 29.32\%             & 42.41\%            & 12.39\%         & 23.08\%         & 11.80\%           & 22.42\%           & \textbf{8.61\%}         & \textbf{9.11\%}        \\
			& PGDAT~\cite{PGD}                   & 86.48\%             & 91.62\%            & 56.44\%         & 69.82\%         & 52.63\%           & 65.38\%           & 71.04\%        & 74.82\%       \\
			& PGDAT+CutOut~\cite{duaninequality}            & 86.92\%             & 91.68\%            & 55.79\%         & 69.79\%         & 52.76\%           & 65.22\%           & 72.58\%        & 76.82\%       \\
			& IGD ($\lambda$=1)        & 65.29\%             & 72.81\%            & 28.76\%         & 45.46\%         & 25.94\%           & 39.91\%           & 61.64\%        & 63.84\%       \\
			& IGD ($\lambda$=2)        & 23.21\%             & 37.41\%            & 10.22\%         & 22.35\%         & 7.76\%            & 18.21\%           & 17.88\%        & 22.02\%       \\
			& IGD ($\lambda$=3)        & 14.86\%             & 27.94\%            & 8.05\%          & \textbf{17.72\%}         & 6.44\%            & \textbf{15.29\%}           & 11.47\%        & 17.16\%       \\
			& IGD ($\lambda$=4)        & \textbf{13.25\%}             & \textbf{25.67\%}            & \textbf{7.00\%}          & 18.41\%         & \textbf{5.69\%}            & 15.75\%           & 11.08\%        & 17.09\%      \\ \bottomrule
		\end{tabular}
	\end{table*}
	\begin{table*}[!t]
		\centering
		\caption{Error rates $\downarrow$ of models across different methods, datasets, and types of IOA guided by different attribution methods. The lowest error rate in each column is \textbf{bold}.}\label{tab:otherIOA}
		\resizebox{\linewidth}{!}{
			\begin{tabular}{@{}cccccccccccccc@{}}
				\toprule
				\multirow{2}{*}{Dataset}      & \multirow{2}{*}{Method} & \multicolumn{3}{c}{Integrated Gradients}               & \multicolumn{3}{c}{Shapley Value}                      & \multicolumn{3}{c}{Input X Gradients}                  & \multicolumn{3}{c}{SmoothGrad}                         \\ \cmidrule(l){3-14} 
				&                         & IOA-B            & IOA-G            & IOA-W            & IOA-B            & IOA-G            & IOA-W            & IOA-B            & IOA-G            & IOA-W            & IOA-B            & IOA-G            & IOA-W            \\ \midrule
				\multirow{7}{*}{CIFAR100}     & Standard                & 72.51\%          & 48.03\%          & 62.77\%          & 61.59\%          & 34.57\%          & 44.52\%          & 59.92\%          & 32.41\%          & 42.33\%          & 48.96\%          & 31.46\%          & 47.84\%          \\
				& PGDAT~\cite{PGD}                  & 59.02\%          & 37.08\%          & 56.51\%          & 37.99\%          & 16.93\%          & 30.77\%          & 40.88\%          & 18.78\%          & 36.70\%          & 37.30\%          & 21.91\%          & 43.28\%          \\
				& PGDAT+CutOut~\cite{duaninequality}           & \textbf{16.90\%} & \textbf{34.66\%} & 56.55\%          & \textbf{12.37\%} & 18.45\%          & 30.22\%          & \textbf{12.68\%} & 18.23\%          & 35.54\%          & \textbf{11.30\%} & 20.18\%          & 41.69\%          \\
				& IGD ($\lambda$=1)       & 57.08\%          & 35.80\%          & 57.53\%          & 35.54\%          & 16.38\%          & 31.03\%          & 38.60\%          & 17.07\%          & 37.13\%          & 36.35\%          & 19.16\%          & 41.57\%          \\
				& IGD ($\lambda$=2)       & 56.74\%          & 35.28\%          & 55.03\%          & 34.12\%          & \textbf{15.27\%} & 29.91\%          & 37.13\%          & 17.09\%          & 35.42\%          & 34.07\%          & \textbf{18.71\%} & 41.69\%          \\
				& IGD ($\lambda$=3)       & 57.74\%          & 35.02\%          & \textbf{52.04\%} & 34.73\%          & 16.74\%          & 29.27\%          & 37.73\%          & \textbf{16.81\%} & \textbf{35.04\%} & 36.04\%          & 19.42\%          & 41.64\%          \\
				& IGD ($\lambda$=4)       & 60.80\%          & 34.95\%          & 53.54\%          & 36.35\%          & 16.14\%          & \textbf{28.40\%} & 39.79\%          & 17.69\%          & 35.09\%          & 39.55\%          & 18.92\%          & \textbf{40.67\%} \\ \midrule
				\multirow{7}{*}{ImageNet-100} & Standard                & \textbf{29.13\%}          & 13.84\%          & \textbf{23.27\%} & 20.32\%          & 7.59\%           & \textbf{13.48\%} & 19.00\%          & 6.67\%           & \textbf{11.37\%} & \textbf{9.63\%}  & 4.83\%           & \textbf{8.74\%}  \\
				& PGDAT~\cite{PGD}                  & 43.49\%          & 14.89\%          & 47.93\%          & 27.65\%          & 10.65\%          & 36.09\%          & 30.83\%          & 11.18\%          & 38.49\%          & 21.30\%          & 8.78\%           & 29.62\%          \\
				& PGDAT+CutOut~\cite{duaninequality}           & 31.07\% & 14.53\%          & 47.73\%          & 20.45\%          & 10.45\%          & 36.49\%          & 21.14\%          & 10.65\%          & 37.34\%          & 15.98\%          & 8.45\%           & 30.54\%          \\
				& IGD ($\lambda$=1)       & 38.86\%          & 17.03\%          & 45.10\%          & 29.32\%          & 10.12\%          & 34.58\%          & 31.03\%          & 10.55\%          & 36.42\%          & 22.29\%          & 9.70\%           & 30.28\%          \\
				& IGD ($\lambda$=2)       & 34.88\%          & 11.97\%          & 36.42\%          & 20.02\%          & 6.18\%           & 24.98\%          & 21.89\%          & 6.15\%           & 26.66\%          & 16.21\%          & 5.62\%           & 22.42\%          \\
				& IGD ($\lambda$=3)       & 35.24\%          & \textbf{10.78\%} & 32.48\%          & 17.72\%          & \textbf{5.39\%}  & 22.65\%          & 19.26\%          & 5.59\%           & 23.11\%          & 14.79\%          & \textbf{3.98\%}  & 19.00\%          \\
				& IGD ($\lambda$=4)       & 31.92\%          & 11.14\%          & 33.53\%          & \textbf{16.50\%} & 5.75\%           & 21.60\%          & \textbf{17.42\%} & \textbf{5.26\%}  & 23.54\%          & 14.23\%          & 4.24\%           & 18.05\%          \\ \bottomrule
			\end{tabular}
		}
	\end{table*}
	
	\subsection{Ablation Study}
	\label{sec:ablation}
	\subsubsection{Other Attribution Methods} \label{sec:otherAM}
	We report the models' inequality-based robustness and the Gini value evaluated by different attribution methods to empirically prove that our claims and findings in Section~\ref{sec:Experiment} generalize to other attribution methods. Results are shown in Table~\ref{tab:otherGiniTable}, Table~\ref{tab:otherINA}, and Table~\ref{tab:otherIOA}.
	
	We can find that the inequality phenomenon and its threats are still not severe on CIFAR100, even with different attribution methods. The gap of $\Gini(A^f(x))$ is small, and the standard-trained model's inequality-based robustness is still worse than the PGDAT-trained model's. On ImageNet-100, results based on Integrated Gradients~\cite{IG}, Input X Gradient~\cite{inputx}, and Shapley Value~\cite{gradShap} are similar to that based on Saliency Map~\cite{saliency}. These results are consistent with our findings in Section~\ref{sec:releaseInequality} to Section~\ref{sec:occRobustnessResult}.
	
	One interesting finding in Table~\ref{tab:otherGiniTable} is that IGD has higher $\Gini(A_{r}^{f}(x))$ if the attribution map is generated by SmoothGrad. We think this is because SmoothGrad adds Gaussian noise to the image before generating the attribution map. In Figure~\ref{fig:imagenet100Visual}\subref{visSmooth}, we can find that the warm color pixels in the attribution map of PGDAT tend to scatter in the background while that of IGD tend to concentrate on the object. This finding suggests that the IGD-trained model can still pay attention to the object even if the image is perturbed by noise, which may empirically explain IGD's better robustness against noise.
	
	Another interesting finding is in Table~\ref{tab:otherINA}, where the standard-trained model has the best INA-robustness if SmoothGrad guides the INA. This is because SmoothGrad, which calculates a local average
	of input gradient's values, can generate an attribution map that is less noisy and more meaningful~\cite{smoothGrad}. SmoothGrad may help INA choose the pixels that are truly important to perturb. We believe SmoothGrad can better expose the threats of the inequality phenomenon and reveal the better inequality-based robustness that a model with lower $\Gini(A^{f}(x))$ truly has.
	
	To sum up, our findings and claims in previous sections still hold even with different attribution methods, and SmoothGrad~\cite{smoothGrad} can better expose the inequality phenomenon and its threats.
	
	\begin{table*}[!t]
		\centering
		\caption{Standard accuracy, adversarial accuracy, global Gini value, regional Gini value, $\left\|\frac{\partial f^{y}(x)}{\partial x}\right\|_{1}$, and the error rates facing different types of IOA. The model is ResNet18. The datasets is ImageNet-100.}
		\begin{tabular}{@{}ccccccccc@{}}
			\toprule
			Method                 & Std. Acc $\uparrow$ & Adv. Acc $\uparrow$ & $\Gini(A^{f}(x))$  & $\Gini(A_{r}^{f}(x))$ & $\left\|\frac{\partial f^{y}(x)}{\partial x}\right\|_{1}$      & IOA-B $\downarrow$   & IOA-G $\downarrow$   & IOA-W $\downarrow$   \\ \midrule
			Standard               & 87.32\%  & 0.00\%      & 0.544 & 0.328 & 4596.42 & 17.59\% & 8.25\%  & 13.71\% \\
			TRADES~\cite{TRADES}                 & 66.98\%  & 30.28\%  & 0.908 & 0.508 & 37.74   & 33.32\% & 10.66\% & 38.48\% \\
			TRADES+CutOut~\cite{duaninequality}          & 66.70\%  & 29.26\%  & 0.906 & 0.509 & 38.61   & 27.65\% & 9.90\%  & 37.68\% \\ \midrule
			TRADES+IGD($\lambda$=1) & 67.58\%  & 28.12\%  & 0.741 & 0.508 & 36.60   & 23.81\% & 6.92\%  & 31.00\% \\
			TRADES+IGD($\lambda$=2) & 69.52\%  & 28.94\%  & 0.686 & 0.493 & 37.86   & 21.28\% & 4.22\%  & 22.42\% \\
			TRADES+IGD($\lambda$=3) & 68.58\%  & 26.64\%  & 0.651 & 0.468 & 37.90   & 21.31\% & 5.22\%  & 23.11\% \\
			TRADES+IGD($\lambda$=4) & 68.02\%  & 25.14\%  & 0.634 & 0.459 & 37.59   & 17.20\% & 5.50\%  & 25.33\% \\ \bottomrule
		\end{tabular}
		
		\label{tab:tradesTable}
	\end{table*}
	
	\begin{figure*}[!t]
		\begin{minipage}{\linewidth}
			\centering
			\includegraphics[width=0.8\linewidth]{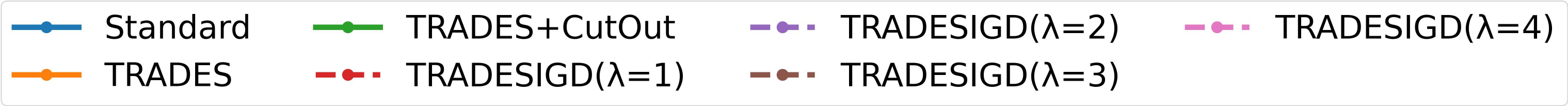}\\
			\subfloat[INA1 on ImageNet-100]{
				\label{imagenet100INA1TRADES}
				\includegraphics[width=0.32\linewidth]{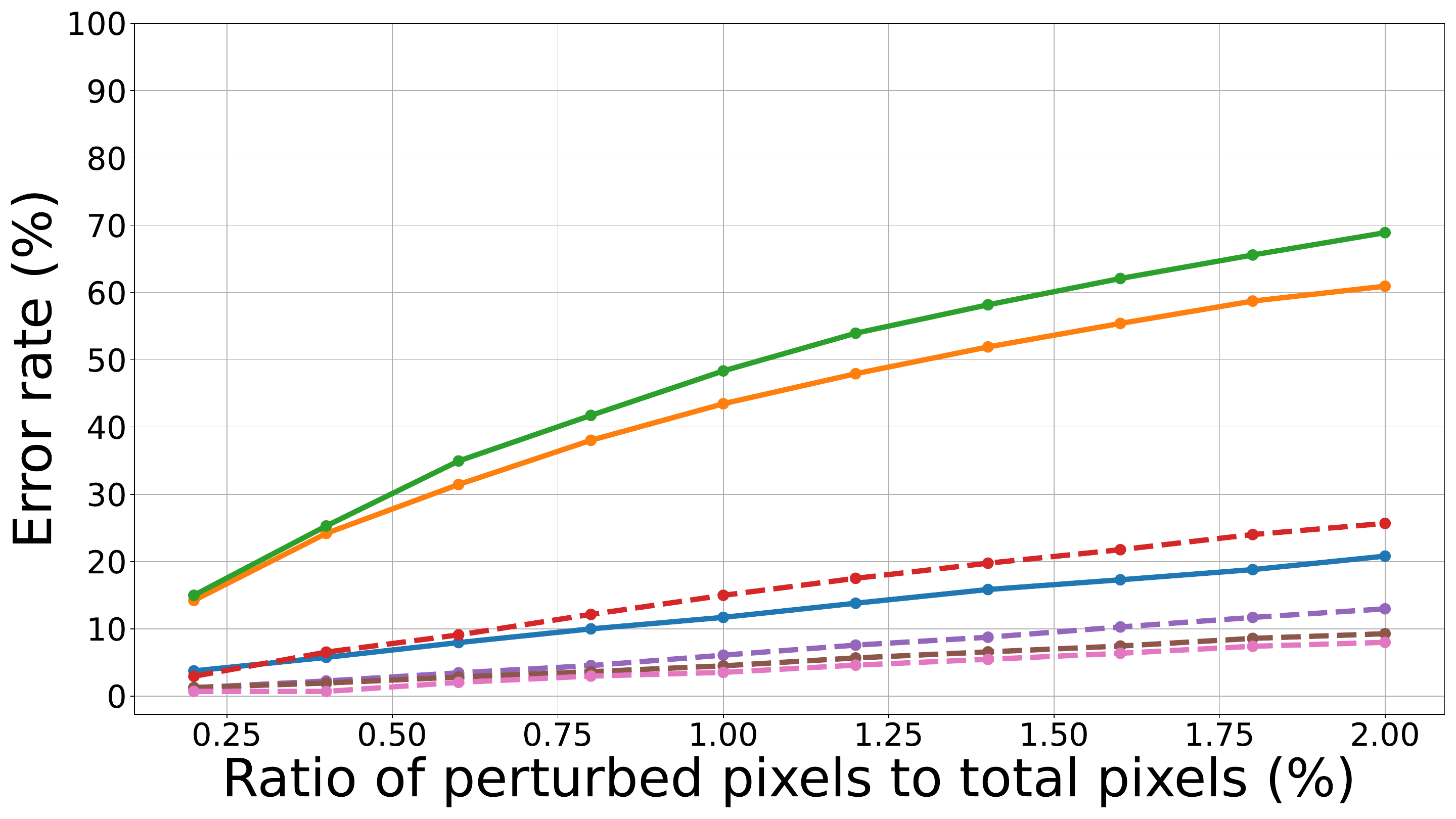}
				
			}
			\subfloat[INA2 on ImageNet-100]{
				\label{imagenet100INA2TRADES}
				\includegraphics[width=0.32\linewidth]{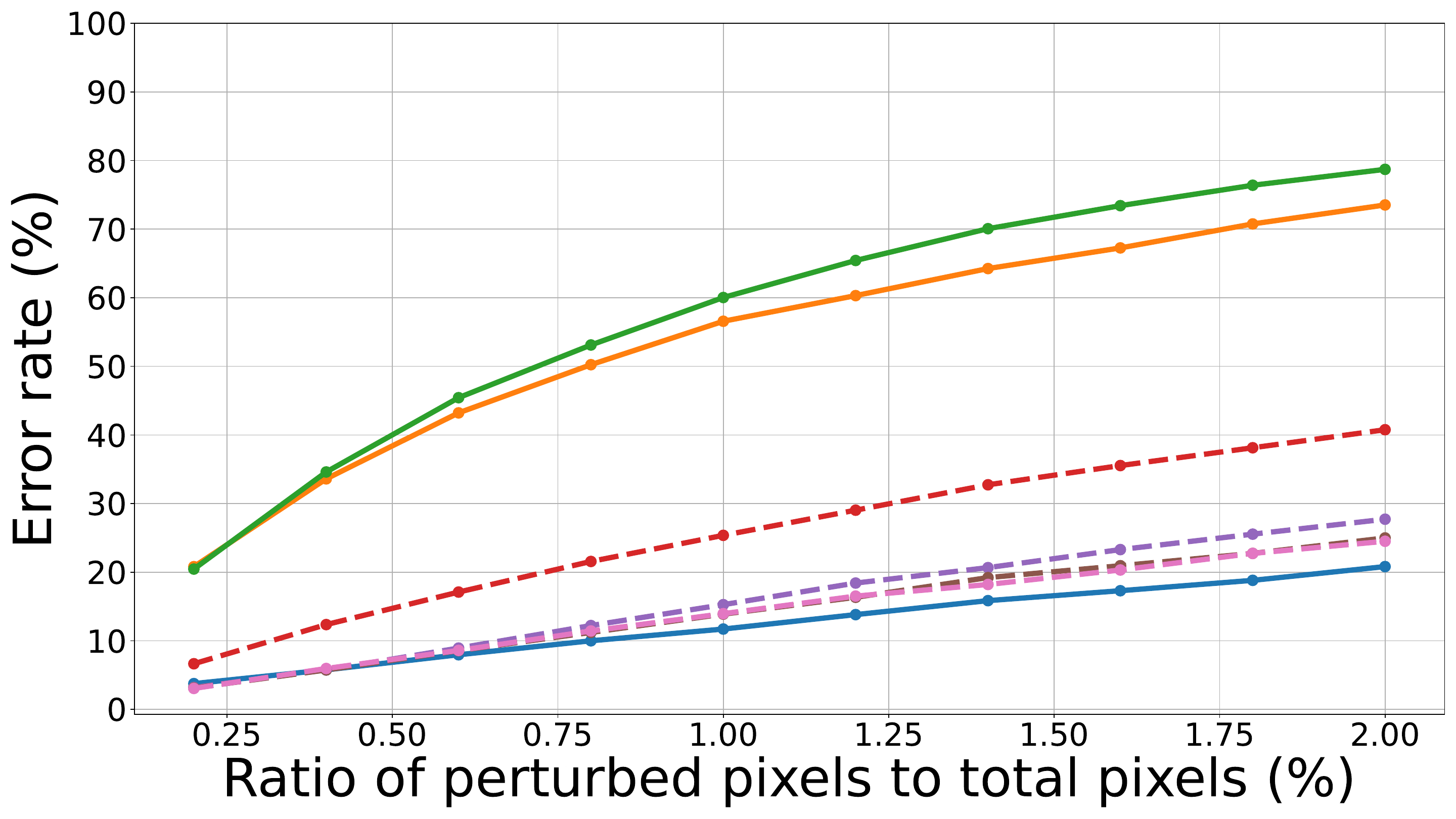}
				
			}
			\subfloat[RN on ImageNet-100]{
				\label{imagenet100RNTRADES}
				\includegraphics[width=0.32\linewidth]{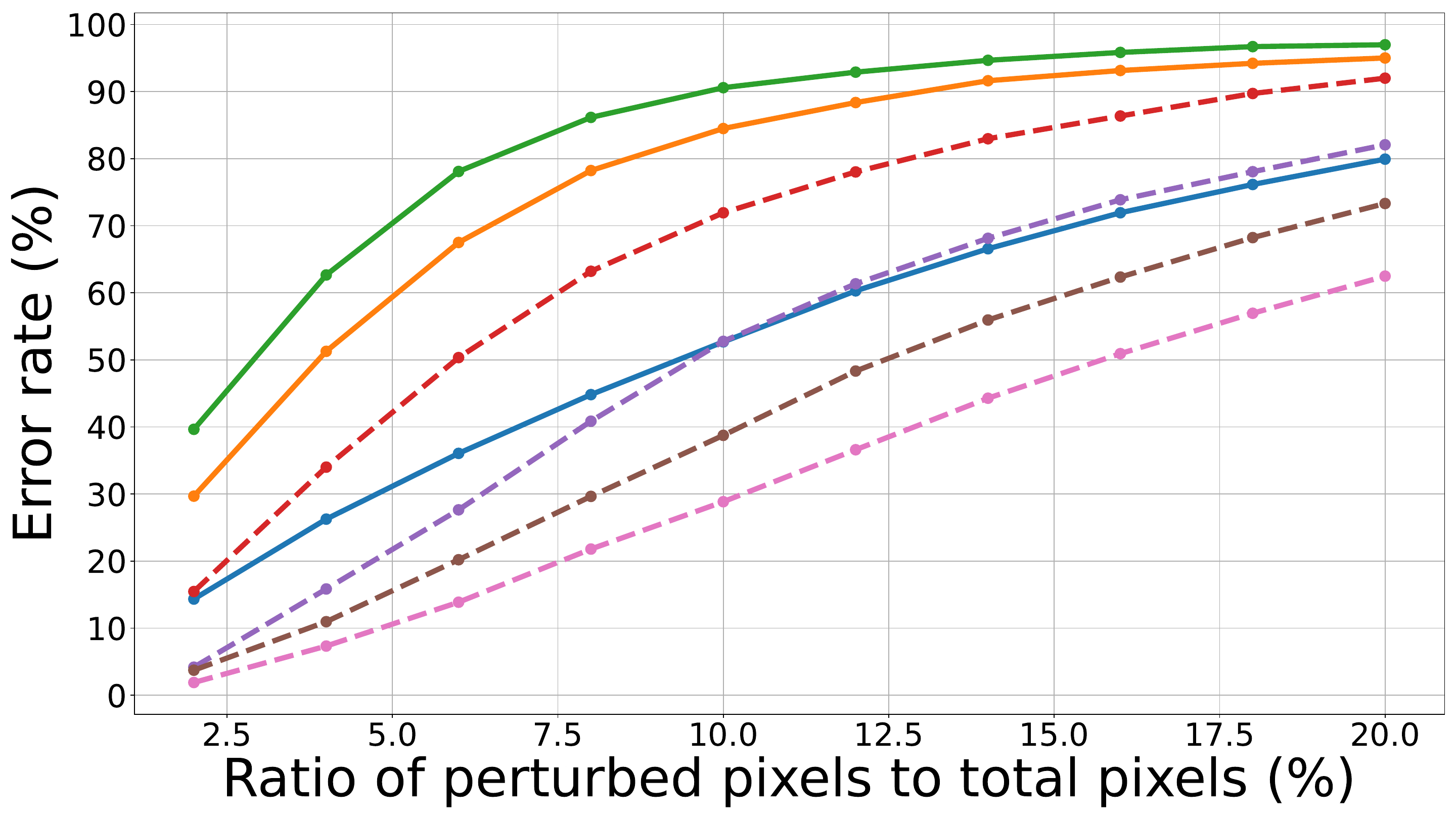}
				
			}
			\caption{Error rates $\downarrow$ of different methods against INA and RN on ImageNet-100. Best viewed in color.}\label{fig:allNoiseTRADES}
		\end{minipage}
	\end{figure*}
	
	\subsubsection{Other Adversarial Training Methods} \label{sec:otherAT}
	
	We also replace the underlying PGDAT of IGD with TRADES~\cite{TRADES}, an adversarial training claimed to be superior to PGDAT, to explore the universality of the inequality phenomenon and IGD's compatibility. Unless otherwise specified, following \citet{TRADES}, we set the regularization terms of TRADES to $6.0$. As shown in \cref{tab:tradesTable}, TRADES also suffer from the inequality phenomenon like the PGDAT, reaching a global Gini Value of 0.906. Consistent with the observations on PGDAT, IGD can still notably reduce the Gini Value of TRADES, demonstrating good compatibility with various adversarial training methods. Concerning the inequality-based robustness, in \cref{tab:tradesTable} and \cref{fig:allNoiseTRADES}, we can find that IGD can consistently improve TRADES's inequality-based robustness.

	\begin{figure}[!t]
		\centering
		\includegraphics[width=\linewidth]{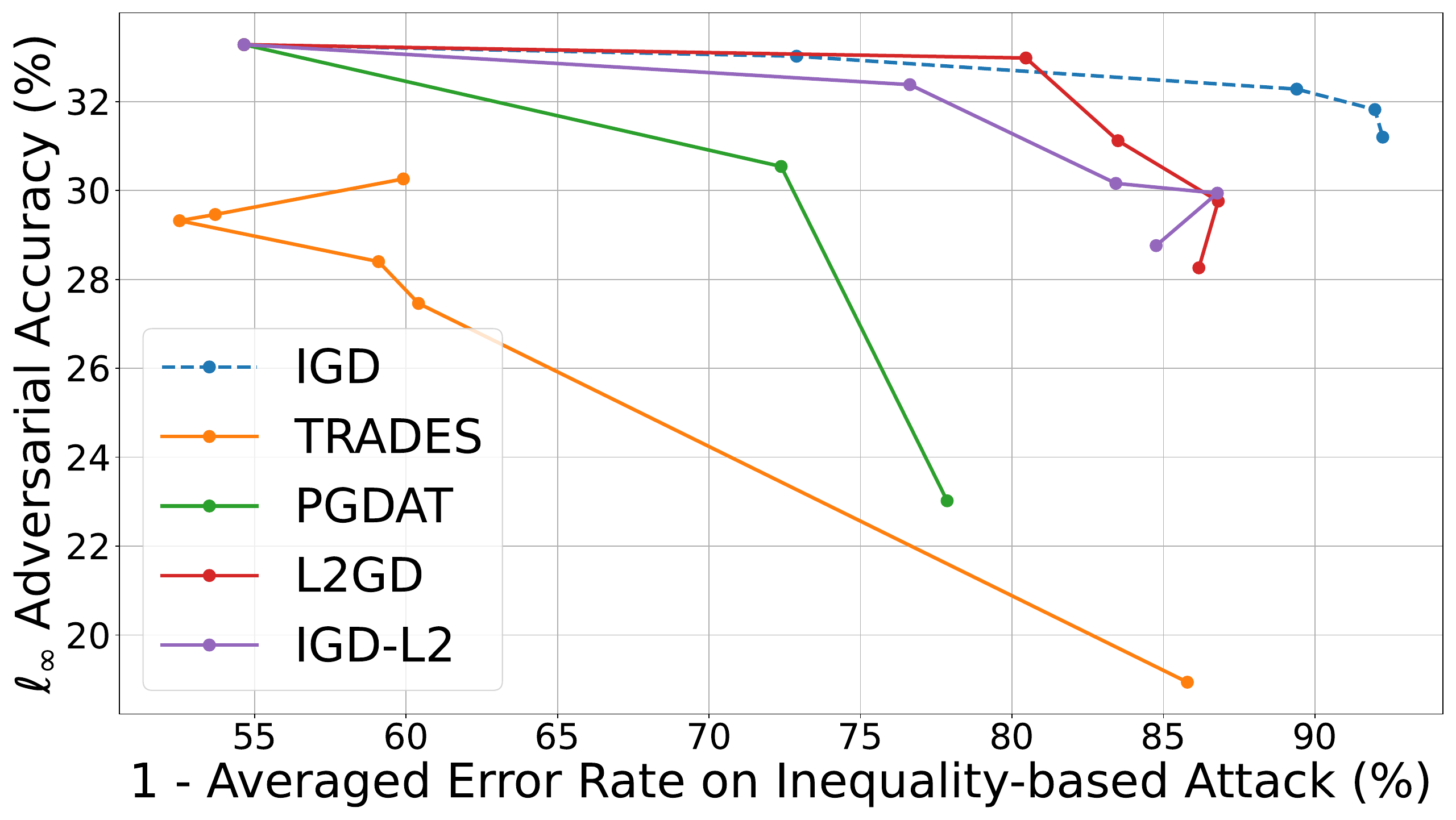}
		\caption{The trade-offs between the $\ell_{\infty}$ adversarial robustness and the inequality-based robustness. Lines and dots on the top-right indicate better trade-offs. Best viewed in color.}
		\label{fig:tradeoff}
	\end{figure}

	\subsubsection{Trade-off between the $\ell_{\infty}$ adversarial robustness and the inequality-based robustness}
	\label{sec:discussTradeoff}

	In Section~\ref{sec:Experiment}, we find that IGD gains inequality-based robustness with 1 to 2\% declines in $\ell_{\infty}$ adversarial robustness. This result may raise a question of whether one can gain remarkable inequality-based robustness by trivially decreasing $\ell_{\infty}$ adversarial robustness.
	
	To answer it, on ImageNet-100, we compare PGDAT~\cite{PGD} with different $\epsilon$ and TRADES~\cite{TRADES} with different regularization terms to IGD. We use averaged error rates on INA, RN, and IOA to represent the inequality-based robustness, where 1\% and 2\% of pixels are perturbed by INA and RN, respectively. Results are presented in Figure~\ref{fig:tradeoff}. We find that one can improve the model's inequality-based robustness by trivially sacrificing the $\ell_{\infty}$ adversarial robustness, i.e., decreasing $\epsilon$ of PGDAT~\cite{PGD} or $\lambda$ of TRADES~\cite{TRADES}. However, to gain inequality-based robustness, these methods have to decline the model's $\ell_{\infty}$ adversarial robustness dramatically and may not be appealing to the one who takes the  $\ell_{\infty}$ adversarial robustness seriously. Furthermore, we find that sacrificing the $\ell_{\infty}$ adversarial robustness by decreasing $\lambda$ of TRADES does not guarantee better inequality-based robustness. These results and findings suggest that gaining acceptable inequality-based robustness for the $\ell_{\infty}$ adversarially trained model is not a trivial task that can be done by simply sacrificing the $\ell_{\infty}$ adversarial robustness.

	\begin{table}[!t]
		\centering
		\caption{Standard accuracy, adversarial accuracy, global Gini value, and regional Gini value of L2GD and IGD-L2. The model is ResNet18. The datasets is ImageNet-100.}
		\resizebox{\linewidth}{!}{\Huge
			\begin{tabular}{@{}cccccc@{}}
				\toprule
				Method               & Std. Acc & Adv. Acc & $\Gini(A^{f}(x))$            & $\Gini(A_{r}^{f}(x))$ & Confidence  \\ \midrule
				IGD-L2($\lambda$=1000)  & 70.40\% & 32.38\% & 0.820 & 0.541 & 70.36\% \\
				IGD-L2($\lambda$=5000)  & 67.92\% & 30.16\% & 0.750 & 0.519 & 67.51\% \\
				IGD-L2($\lambda$=7500)  & 67.50\% & 29.94\% & 0.702 & 0.512 & 66.33\% \\
				IGD-L2($\lambda$=10000) & 66.30\% & 28.76\% & 0.712 & 0.502 & 64.47\% \\\midrule
				L2GD ($\lambda$=1000) & 71.32\%  & 32.98\%  & 0.818 & 0.549 & 70.73\% \\
				L2GD ($\lambda$=2500) & 70.60\%  & 31.12\%  & 0.757 & 0.534 & 70.49\% \\
				L2GD ($\lambda$=5000) & 70.84\%  & 29.76\%  & 0.746 & 0.512 & 70.09\%\\
				L2GD ($\lambda$=7500) & 69.42\%  & 28.26\%  & 0.714 & 0.505 &  69.13\%\\ \bottomrule
				
		\end{tabular}}
		
		\label{tab:L2}
	\end{table}

	\subsubsection{Using $L_2$ Distance for Alignment is Sub-optimal}\label{sec:igdl2}
	
	One key component of our IGD is the Cosine Similarity used to align the input gradients. We also try using $\ell_2$ distance for alignment and name this variant IGD-L2. As presented in \cref{tab:L2} and \cref{fig:tradeoff}, IGD-L2 has far worse standard accuracy, adversarial robustness, and adversarial-inequality trade-off than IGD. Recall \cref{sec:motivation} that the motivation behind using the Cosine Similarity, which normalizes the input gradients, is to prevent the alternation in the norm of the input gradients from degrading the adversarial robustness. Also, as presented in \cref{tab:advGinitable}, the input gradients' norm of standard-trained models is much higher than that of adversarially trained models, indicating that $\ell_{\infty}$-AT will strongly penalize the norm of the input gradients. Thus, aligning the input gradients between the standard- and adversarially trained models without normalization will degrade the adversarially trained model more severely.
	
	\subsubsection{Applying Our Theoretical Conclusion to Discard the Teacher Model}\label{sec:l2gd}
	
	Another key component of our IGD is the standard-trained teacher model, which provides equal input gradients to guide the adversarially trained student model. Yet, The necessity of incurring additional overhead\footnote{A 10-step PGDAT takes 11 forwards and 11 backward for each sample within an epoch and has a total of 150 epochs in this paper. Training the standard-trained model takes one forward and one backward for each sample within an epoch. Considering that the input gradients can be pre-computed, the budget of acquiring the input gradients equals training another extra epoch. Thus, IGD incurs an additional budget of around 9\%. If the standard-trained model is off-the-shelf, which is a common scenario, then the extra budget is around 0.06\%.} for obtaining the standard-trained model and its input gradients remains unproven. That is, can we discard the teacher model and directly penalize the input gradients?
	
	We first tried (1) directly penalizing the Gini Value of the input gradients and (2) setting the target of \cref{eq:alignLoss} to $\text{sign}(\frac{\partial f_{\theta_{Adv}}^{y}(x)}{\partial x})$, whose Gini Value is 0. However, through massive hyper-parameter tuning, we find that both two methods can not train usable models and achieve only around 2\% accuracy on ImageNet-100. It seems that directly equalizing the input gradients will disturb the training.
	
	Fortunately, recalling \cref{eq:deltaVar} that the squared norm of the input gradients has the same monotonicity as the Gini value, we can derive a surrogate loss that equalizes the input gradients by penalizing the squared norm of the input gradients. That is, we minimize
	\begin{equation}
		\mathcal{L} = \mathcal{L}_{ce} + \lambda * \left\|\frac{\partial f_{\theta_{Adv}}^{y}(x)}{\partial x}\right\|_{2}^{2}\,,
	\end{equation}
	named $\ell_{2}$-norm gradient decay (\textbf{L2GD}). In \cref{tab:L2}, we observe that L2GD can also release the inequality phenomenon, demonstrating that our theoretical results are applicable. However, L2GD suffers from a more severe standard and adversarial accuracy drop than IGD. In \cref{fig:tradeoff}, we can find that IGD still achieves similar and mostly better trade-offs than L2GD and that IGD can achieve better inequality-based robustness than L2GD (92.24\% vs 86.17\%) by increasing $\lambda$ before it saturates. We hypothesize the reason is that the standard-trained model's input gradients used by IGD contain information on classification, which can better assist the student model to gain a good performance (e.g., higher classification confidence).
	
	To conclude, by applying our theoretical conclusion, releasing the inequality phenomenon with a simple strategy that does not rely on the standard-trained model is feasible but currently can not achieve better performance than IGD.

	\section{Conclusion}
	\label{sec:Conclusion}
	
	In this paper, we propose a simple yet effective method called Input Gradient Distillation (IGD) to release the inequality phenomenon in $\ell_{\infty}$-AT. While preserving the model's adversarial robustness, IGD can make the $\ell_{\infty}$-adversarially trained model's decision pattern equal and improve the $\ell_{\infty}$-adversarially trained model's robustness against inductive attacks~\cite{duaninequality} and i.i.d. random noise. Besides releasing the inequality phenomenon, IGD is also a tool to help us analyze and explain the relationship between the inequality phenomenon and the model's robustness against noise and occlusion because of its ability to control variables. With the help of IGD, we point out that releasing the inequality phenomenon can decrease the class score's deviation caused by the input space perturbation. Moreover, we discover that the severity of the inequality phenomenon may vary according to the resolution of the dataset and that the inequality phenomenon is more severe on the high-resolution dataset. Through empirical experiments, we conjecture that such a phenomenon is caused by the intrinsic properties of the low-resolution dataset and the influence of other factors that may overwhelm the inequality phenomenon when the resolution of the dataset is low.
	
	We hope that this work can inspire future research to propose better methods for resolving and evaluating the threats of the inequality phenomenon and to provide more insightful findings and explanations of how the inequality phenomenon couples with other factors, like the resolution of the dataset we discuss, in terms of the model's robustness. Furthermore, other than $\ell_{\infty}$-AT, a thorough investigation evaluating the inequality phenomenon on various machine learning techniques that have been proposed can be appealing since this can check whether the inequality phenomenon is common and can help the community to have a more comprehensive understanding of the inequality phenomenon in the field of machine learning.

	{
		\footnotesize
		\bibliographystyle{IEEEtranN}
		\bibliography{main}
	}

	\begin{IEEEbiography}[{\includegraphics[width=1in,height=1.25in,clip,keepaspectratio]{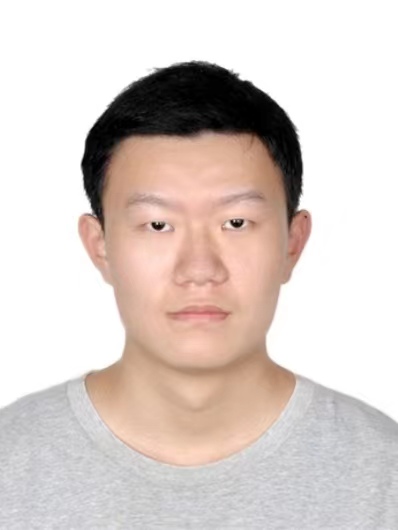}}]{Junxi Chen}
	received the B.S. degree in computer science and technology from Sun Yat-sen University, Guangzhou, China, in 2023. He is currently pursuing the Ph.D. degree in computer science and technology from Sun Yat-sen University, Guangzhou, China. His research interest is adversarial machine learning.
\end{IEEEbiography}
\begin{IEEEbiography}[{\includegraphics[width=1in,height=1.25in,clip,keepaspectratio]{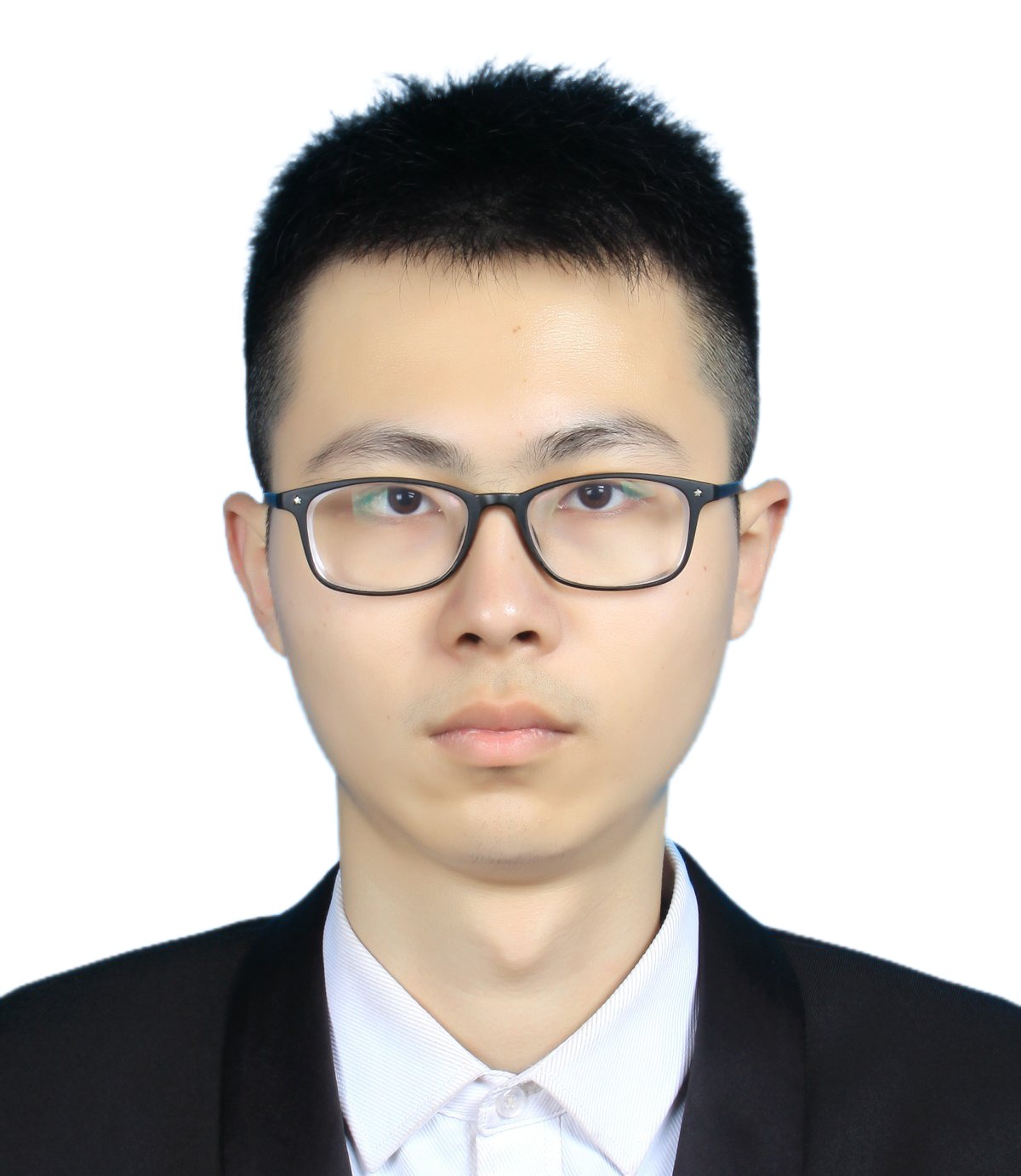}}]{Junhao Dong}
	received the M.S. degree in computer science and technology from Sun Yat-sen University, Guangzhou, China, in 2023. He is currently working toward the Ph.D. degree with the College of Computing and Data Science, Nanyang Technological University, Singapore. His research interests include trustworthy AI, computer vision, and adversarial machine learning.
\end{IEEEbiography}
\begin{IEEEbiography}[{\includegraphics[width=1in,height=1.25in,clip,keepaspectratio]{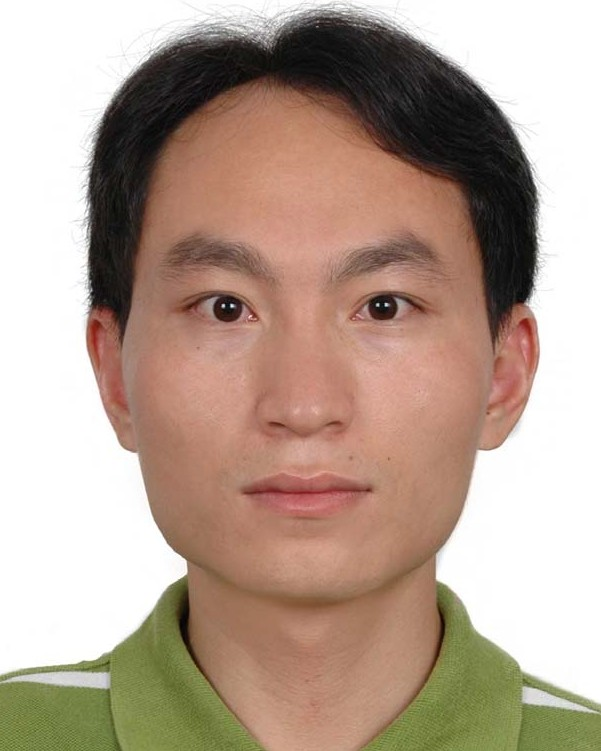}}]{Xiaohua Xie}
	(Member, IEEE) received the Ph.D. degree in applied mathematics from Sun Yat-sen University, China, in 2010. He was an Associate Professor with the Shenzhen Institutes of Advanced Technology (SIAT), Chinese Academy of Sciences. He is currently an Associate Professor with Sun Yat-sen University. His current research interests include cover image processing, computer vision, pattern recognition, and machine learning.
\end{IEEEbiography}
\begin{IEEEbiography}[{\includegraphics[width=1in,height=1.25in,clip,keepaspectratio]{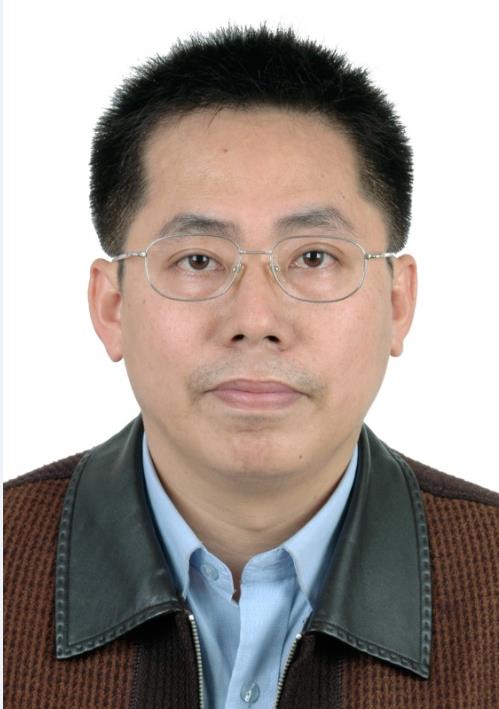}}]{Jianhuang Lai}
	(Senior Member, IEEE) received the Ph.D. degree in mathematics from Sun Yat-sen University, China, in 1999. In 1989, he joined Sun Yat-sen University as an Assistant Professor, where he is currently Professor with the School of Computer science and Engineering. His current research interests include the areas of computer vision, pattern recognition, and its applications. He has published over 250 scientific papers in the international journals and conferences on image processing and pattern recognition. He serves as the Deputy Director of the Image and Graphics Association of China.
\end{IEEEbiography}

\end{document}